\documentclass{article}

\usepackage[preprint,nonatbib]{neurips_data_2024}

\usepackage[utf8]{inputenc} 
\usepackage[T1]{fontenc}    
\usepackage{hyperref}       
\usepackage{url}            
\usepackage{booktabs}       
\usepackage{amsfonts}       
\usepackage{amsmath}
\usepackage{amsthm}
\usepackage{nicefrac}       
\usepackage{microtype}      
\usepackage{xcolor}         

\usepackage{adjustbox}
\usepackage{caption}
\usepackage{makecell}
\usepackage{subcaption}
\usepackage{comment}
\usepackage{multirow}

\usepackage[backend=biber]{biblatex}
\DeclareMathOperator*{\argmin}{arg\,min}

\theoremstyle{plain}
\newtheorem{theorem}{Theorem}[section]

\title{HoTPP Benchmark: Are We Good at the Long Horizon Events Forecasting?}

\author{%
  Ivan Karpukhin \\
  Sber AI Lab\\
  \texttt{iakarpukhin@sberbank.ru} \\
  \And
  Foma Shipilov \\
  Skoltech, Sber AI Lab \\
  \texttt{foma.shipilov@skoltech.ru} \\
  \And
  Andrey Savchenko \\
  Sber AI Lab \\
  \texttt{avladsavchenko@sberbank.ru} \\
}

\begin{document}

\maketitle

\begin{abstract}
Forecasting multiple future events within a given time horizon is essential for applications in finance, retail, social networks, and healthcare. Marked Temporal Point Processes (MTPP) provide a principled framework to model both the timing and labels of events. However, most existing research focuses on predicting only the next event, leaving long-horizon forecasting largely underexplored. To address this gap, we introduce \textit{HoTPP}, the first benchmark specifically designed to rigorously evaluate long-horizon predictions. We identify shortcomings in widely used evaluation metrics, propose a theoretically grounded \textit{T-mAP} metric, present strong statistical baselines, and offer efficient implementations of popular models. Our empirical results demonstrate that modern MTPP approaches often underperform simple statistical baselines. Furthermore, we analyze the diversity of predicted sequences and find that most methods exhibit mode collapse. Finally, we analyze the impact of autoregression and intensity-based losses on prediction quality, and outline promising directions for future research. The HoTPP source code, hyperparameters, and full evaluation results are available at \url{https://github.com/ivan-chai/hotpp-benchmark}.
\end{abstract}

\section{Introduction}
Internet activity, e-commerce transactions, retail operations, clinical visits, and numerous other aspects of our lives generate vast amounts of data in the form of timestamps and related information. In the era of AI, it is crucial to develop methods capable of handling these complex data streams. We refer to this type of data as Event Sequences (ESs).

Event sequences differ fundamentally from other data types. Unlike tabular data~\cite{wang2022transtab}, ESs include timestamps and possess an inherent order. In contrast to time series data~\cite{lim2021timeseriessurvey}, ESs are characterized by irregular time intervals and additional data fields, such as event types. These differences necessitate the development of specialized models and evaluation practices.

Irregularly spaced timestamps are commonly modeled via Temporal Point Processes (TPP) \cite{rizoiu2017hawkes-tutorial}. Assigning categorical labels to these events leads to the Marked Temporal Point Process (MTPP) formalism. Since most event-sequence data can be viewed as MTPPs augmented with additional fields \cite{mcdermott2024eventstreamgpt}, accurately modeling event times and labels remains a central challenge in the broader event-stream (ES) domain.

In practice, a common question arises: what events will occur, and when, within a specific \textit{time horizon}? Forecasting multiple future events presents unique challenges that differ from traditional next-event prediction tasks~\cite{xue2022hypro,zhou2025diffusiontpp}. For instance, autoregressive event sequence prediction involves applying the model to its own potentially erroneous predictions. However, the challenges of autoregressive prediction in the context of MTPP have not been thoroughly explored. Another difficulty lies in evaluation. Methods like Dynamic Time Warping (DTW) are typically unsuitable for event sequences due to strict ordering constraints~\cite{su2017orderwasserstein}. Some works have applied Optimal Transport Distance (OTD), a variant of the Wasserstein distance, to compare sequences of predefined lengths \cite{mei2019imputing,xue2022hypro,xue2023easytpp}. However, the limitations of this metric have not been previously considered.


\begin{figure*}[t]
\centering
  \includegraphics[width=\textwidth]{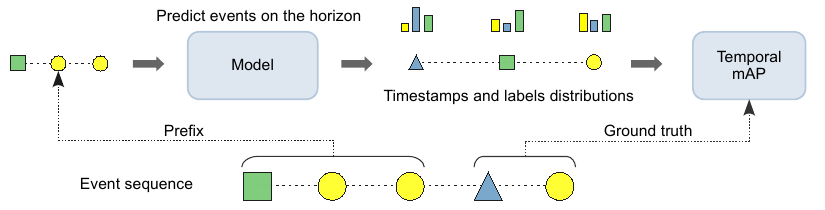}
  \caption{T-mAP evaluation pipeline. Unlike previous approaches, T-mAP evaluates sequences of variable lengths within the prediction horizon. It further improves performance assessment by analyzing label distributions rather than relying on fixed predictions.
  }
  \label{fig:brief}
\end{figure*}

In this work, we provide the first in-depth analysis of models and metrics for long-horizon event forecasting, establishing a rigorous evaluation framework and a baseline for MTPP studies. To this end, we present \textit{HoTPP}, a novel benchmark illustrated in Figure~\ref{fig:brief}. Our key contributions are summarized as follows:
\begin{enumerate}
    \item We introduce the first open-source benchmark specifically designed for long-horizon event forecasting. By considering datasets and methods spanning finance, social networks, healthcare, and recommender systems, HoTPP significantly broadens the diversity and scale of data compared to previous benchmarks~\cite{xue2023easytpp,mcdermott2024eventstreamgpt,xu2018poppy}. Moreover, we provide optimized training and inference algorithms to support large-scale studies.
    \item We demonstrate that widely used evaluation metrics for MTPPs overlook critical aspects of model performance. In particular, they are sensitive to calibration and incorrectly count false positives and false negatives. To address these issues, we introduce Temporal mean Average Precision (T-mAP), a novel evaluation metric inspired by best practices in computer vision~\cite{lin2014mscoco}. Furthermore, we address a theoretical gap in computer vision by proving the correctness of the T-mAP computation algorithm.
    \item Guided by our methodology, we evaluate the forecasting performance of a diverse set of approaches. Our findings indicate that methods excelling at next-event prediction often underperform in long-horizon scenarios, and vice versa. Moreover, in 6 out of 10 long-horizon tasks, simple statistical baselines either outperform or match modern MTPP models, highlighting the need for more effective solutions.
    \item Finally, we quantify the entropy of predicted labels and demonstrate that most modern approaches tend to predict only the most frequent labels, suffering from the problem known as \textit{mode collapse}. This issue severely restricts practical applicability and motivates further research into more accurate prediction strategies.
\end{enumerate}

\section{Related Work}
{\bf Marked Temporal Point Processes.}
MTPP is a stochastic process consisting of a sequence of pairs $(t_1, l_1), (t_2, l_2), \dots$, where $t_1 < t_2 < \dots$ represent event times and $l_i \in \{1, \dots, L\}$ denote event type labels~\cite{rizoiu2017hawkes-tutorial}. Common approaches to MTPP modeling focus on predicting the next event. A basic solution involves independently predicting the event time and type. A more advanced approach splits the original sequence into subsequences, one for each event type, and independently models each subsequence's timing. Depending on the form of time-step distribution, the process is called Poisson or Hawkes.

Over the last decade, the focus has shifted toward increasing model flexibility by applying neural architectures. Several works have employed classical RNNs~\cite{du2016rmtpp,xiao2017rnnintencities,omi2019fullynn} and transformers~\cite{zuo2020thp,zhang2020sahp}, while others have proposed architectures with continuous time~\cite{mei2017nhp,rubanova2019ode,kuleshov2024cotode}. In this work, we evaluate MTPP models with both discrete and continuous time architectures. Unlike previous benchmarks, we also assess simple rule-based baselines and popular methods from related fields, including GPT-like prediction models for event sequences~\cite{mcdermott2024eventstreamgpt,padhi2021tabgpt}. 
For more details on MTPP modeling, refer to \ref{app:background-tpp}.

{\bf MTPP Evaluation.} Previous benchmarks have primarily focused on medical data or traditional MTPP datasets. EventStream-GPT~\cite{mcdermott2024eventstreamgpt} and TemporAI~\cite{saveliev2023temporai} consider only medical data and do not implement methods from the MTPP field, despite their applicability. Early MTPP benchmarks, such as Tick~\cite{bacry2017tick} and PyHawkes\footnote{\url{https://github.com/slinderman/pyhawkes}}, implement classical machine learning approaches but exclude modern neural networks. While PoPPy~\cite{xu2018poppy} and EasyTPP~\cite{xue2023easytpp} include neural methods, they do not consider rule-based and horizon prediction approaches, such as HYPRO~\cite{xue2022hypro} and Diffusion~\cite{zhou2025diffusiontpp}. Furthermore, PoPPy does not evaluate long-horizon predictions at all. EasyTPP addresses this limitation to some extent by reporting the OTD metric, though it does not provide the corresponding evaluation code.

Previous work also highlighted the limitations of Dynamic Time Warping (DTW) for event sequence evaluation~\cite{su2017orderwasserstein}, showing that DTW's strict ordering constraints are impractical and should be avoided. This work proposed an alternative, the Order-preserving Wasserstein Distance (OPW). In our study, we address the limitations of OTD, a variant of the Wasserstein Distance, and propose a new metric called T-mAP. Unlike OPW, T-mAP operates on timestamps rather than event indices and involves two hyperparameters, compared to three in OPW.

\section{T-mAP Metric}
\label{sec:metric}
In this section, we show that widely used evaluation metrics for long-horizon prediction often miscount false positives (FP) and false negatives (FN). We then introduce \textit{Temporal mean Average Precision} (T-mAP), a novel, theoretically grounded metric that addresses these issues. In addition, T-mAP is invariant to linear calibration of model outputs, simplifying and stabilizing comparisons across different methods. A general overview of the T-mAP evaluation pipeline is presented in Fig.~\ref{fig:brief}.

\subsection{Limitations of the Next-Event and OTD Metrics}
\label{sec:otd-limitations}

MTPPs are typically evaluated based on the accuracy of next-event predictions, with time and type predictions assessed independently. The quality of type predictions is measured by the error rate, while time prediction error is evaluated using either Mean Absolute Error (MAE) or Root Mean Squared Error (RMSE). However, these metrics do not account for the model's ability to predict multiple future events. For example, in autoregressive models, the outputs are fed back as inputs for subsequent steps, which can lead to cumulative prediction errors. As a result, long-horizon evaluation metrics are necessary.

\begin{figure*}[t]
\centering
  \includegraphics[width=\textwidth]{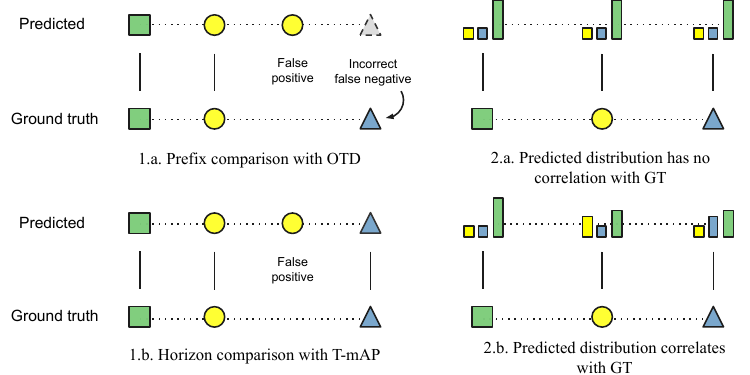}
  \caption{
  Comparison of the OTD and T-mAP metrics. In the example 1.a, OTD’s prefix-based evaluation of three events erroneously produces a false negative, whereas T-mAP in example 1.b compares all events within the specified time horizon for a more accurate assessment. Moreover, OTD evaluates only the label with the highest probability, treating cases 2.a and 2.b as equivalent. By contrast, T-mAP considers the entire distribution, yielding a low score (0.33) for 2.a and a perfect score (1) for 2.b.
  }
  \label{fig:otd-limitations}
\end{figure*}

Recent studies have advanced the measurement of horizon prediction quality by employing OTD~\cite{mei2019imputing}. This metric is analogous to the edit distance between the predicted event sequence and the ground truth. Suppose there is a predicted sequence $S_p = \{(t^p_i, l^p_i)\}_{i=1}^{n_p}, 0 < t_1^p < t_2^p < \dots$, and a ground truth sequence $S_{gt} = \{(t^{gt}_j, l^{gt}_j)\}_{j=1}^{n_{gt}}, 0 < t_1^{gt} < t_2^{gt} < \dots$. These two sequences form a bipartite graph $\mathcal{G}(S_p, S_{gt})$. The prediction is connected to the ground truth event if their types are equal. Denote $M(S_p, S_{gt})$ a set of all possible matches in the graph $\mathcal{G}(S_p, S_{gt})$. OTD finds the minimum cost among all possible matchings:
\begin{equation}
    \mathrm{OTD}(S_p, S_{gt}) = \min\limits_{m \in M(S_p, S_{gt})}\left[ \sum\limits_{(i, j) \in m}|t^p_i - t^{gt}_j| + C_{del} U_p(m) + C_{ins} U_{gt}(m) \right],
\end{equation}
where $U_p(m)$ is the number of unmatched predictions in matching $m$, $U_{gt}(m)$ is the number of unmatched ground truth events, $C_{ins}$ is an insertion cost, and $C_{del}$ is a deletion cost. It is common to take $C_{ins} = C_{del}$ \cite{mei2019imputing}. It has been proven that OTD is a valid metric, as it is symmetric, equals zero for identical sequences, and satisfies the triangle inequality.

OTD is computed over fixed-size prefixes, limiting its flexibility when models produce imprecise time-step predictions, as shown in Fig.~\ref{fig:otd-limitations}.1. When a model predicts events too frequently, the first $K$ predicted events will correspond to the early part of the ground truth sequence, resulting in false negatives. This outcome is misleading because considering a longer predicted sequence would alter the evaluation. Conversely, if the predicted time step is too large, the first $K$ predicted events will extend far beyond the horizon of the first $K$ target events, leading to a significant number of incorrect false positives. 
Therefore, evaluating a dynamic number of events is essential to align with the ground truth's time horizon properly.

Metrics for MTPP evaluation can be categorized into two groups: those relying on event indices and those that measure time and label prediction quality independently of event sequential positions. Next-event metrics depend on ordering, even if multiple events share the same timestamp, which makes the order meaningless. Although OTD is formally invariant to order, extracting sequences for comparison still depends on event indices. For instance, when OTD compares prefixes of length $K$, the first $K$ predicted events must align with the ground truth to minimize the OTD score. As a result, OTD, like next-event metrics, can be influenced by ordering even when it cannot be uniquely determined. This dependency is non-trivial and cannot be easily measured or controlled.

Another limitation of the OTD metric is its inability to evaluate the full predicted distribution of labels, as shown in Fig.~\ref{fig:otd-limitations}.2. OTD considers only the labels with the highest probability, ignoring the complete distribution. This makes it dependent on model calibration and limits the ability to assess performance across a broader range of event types, such as long-tail predictions. However, models typically predict probabilities for all classes, allowing for a more comprehensive assessment. Therefore, we aim to develop an evaluation metric that accurately captures performance across common and rare classes.

\subsection{Temporal mAP: Definition and Computation}
In this section, we introduce a novel metric, {\it Temporal mAP (T-mAP)}, which analyses all errors within a predefined time horizon, explicitly controls ordering, and is invariant to linear calibration. T-mAP is inspired by object detection methods from computer vision~\cite{everingham2010pascalcvdetection}. Object detection aims to localize objects within an image and identify their types. In event sequences, we tackle a similar problem but consider the time dimension instead of horizontal and vertical axes. Unlike object detection, where objects have spatial size, each event in an MTTP is a point without a duration. Therefore, we replace the intersection-over-union (IoU) similarity between bounding boxes with the absolute time difference.


T-mAP incorporates concepts from OTD but addresses its limitations, as outlined in the previous section. Firstly, T-mAP evaluates label probabilities instead of relying on final predictions. Secondly, T-mAP restricts the prediction horizon rather than the number of events. These adjustments result in significant differences in formulation and computation, detailed below.

\subsubsection{Definition}
T-mAP is parameterized by the horizon length $T$ and the maximum allowed time error $\delta$. T-mAP compares predicted and ground truth sequences within the interval $T$ from the last observed event. Consider a simplified scenario with a single event type $l$. Assume the model predicts timestamps and presence scores (logits or probabilities) for several future events: $S^l_p = \{(t^p_i, s^p_i)\}, 1 \le i \le n_p$. The corresponding ground truth sequence is $S^l_{gt} = \{t^{gt}_j\}, 1 \le j \le n_{gt}$. For simplicity, assume the sequences $S^l_p$ and $S^l_{gt}$ are filtered to include only events within the horizon $T$.

For any threshold value $h$, we can select a subset of the predicted sequence $S^l_p$ with scores exceeding the threshold:
\begin{equation}
S^l_>(h) = \{t^p_i : \exists (t^p_i, s^p_i) \in S^l_p, s^p_i > h\}.
\end{equation}
By definition, a predicted event $i$ can be matched with a ground truth event $j$ iff $|t^p_i - t^{gt}_j| \le \delta$, meaning the time difference between the predicted and ground truth events is less than or equal to $\delta$. T-mAP identifies the matching that maximizes precision and recall, i.e., matching with maximum cover $c$. The precision of this matching is calculated as $c / |S^l_>(h)|$ and the recall as $c / |S^l_{gt}|$. For any threshold $h$, we can count true positives (TP), false positives (FP), and false negatives (FN) across all predicted and ground truth sequences. Note that there are no true negatives, as the model cannot explicitly predict the absence of an event. Similar to binary classification, we can define a precision-recall curve by varying the threshold $h$ and then estimate the Average Precision (AP), the area under the precision-recall curve. Finally, T-mAP is defined as the average AP across all classes.

\subsubsection{Computation}
\label{sec:map-computation}
Finding the optimal matching independently for each threshold value $h$ is impractical; thus, we need a more efficient method to evaluate T-mAP. This section shows how to optimize T-mAP computation using an assignment problem solver, like the Jonker-Volgenant algorithm~\cite{jonker1988linsum}. The resulting complexity of T-mAP computation is $\mathcal{O}(B N^3)$, where $B$ is the number of evaluated sequences and $N = \max(n_p, n_{gt})$ is the number of events within the horizon.

\begin{figure*}[t]
\centering
  \includegraphics[width=\textwidth]{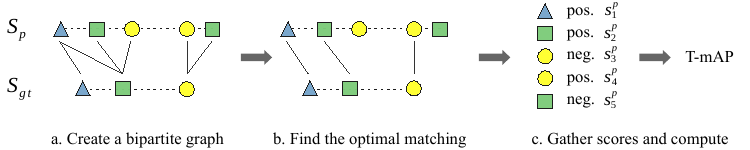}
  \caption{T-mAP computation pipeline.}
  \label{fig:computation}
\end{figure*}

For each pair of sequences $S^l_p$ and $S^l_{gt}$ we define a weighted bipartite graph $\mathcal{G}(S^l_p, S^l_{gt})$ with $|S^l_p|$ vertices in the first and $|S^l_{gt}|$ vertices in the second part. For each pair of prediction $i$ and ground truth event $j$ with $|t^p_i - t^{gt}_j| \le \delta$ we add an edge with weight $-s^p_i$, equal to negative logit of the target class, as shown in Fig.~\ref{fig:computation}.a. Jonker-Volgenant algorithm finds the matching with the maximum number of edges in the graph, such that the resulting matching minimizes the total cost of selected edges, as shown in Fig.~\ref{fig:computation}.b. We call this matching {\it optimal matching} and denote a set of all optimal matching possibilities as $M(\mathcal{G})$. For any threshold $h$, there is a subgraph $\mathcal{G}_h(S^l_>(h), S^l_{gt})$ with events whose scores are greater than $h$. The following theorem holds:
\begin{theorem}
\label{theor:map-computation}
For any threshold $h$ there exists an optimal matching in the graph $\mathcal{G}_h$, that is a subset of an optimal matching in the full graph $\mathcal{G}$:
\begin{equation}
    \forall h \forall m \in M(\mathcal{G}) \exists m_h \in M(\mathcal{G}_h): m_h \subset m.
\end{equation}
\end{theorem}
According to this theorem, we can compute the matching for the prediction $S^l_{gt}$ and subsequently reuse it for all thresholds $h$ and subsequences $S^l_>(h)$ to construct a precision-recall curve for the entire dataset, as shown in Fig.~\ref{fig:computation}.c. The proof of the theorem, the study of calibration dependency, and the complete algorithm for T-mAP evaluation are provided in \ref{app:tmapcomp}.

\subsubsection{T-mAP Hyperparameters}
\begin{figure*}[t]
\centering
  \includegraphics[width=\textwidth]{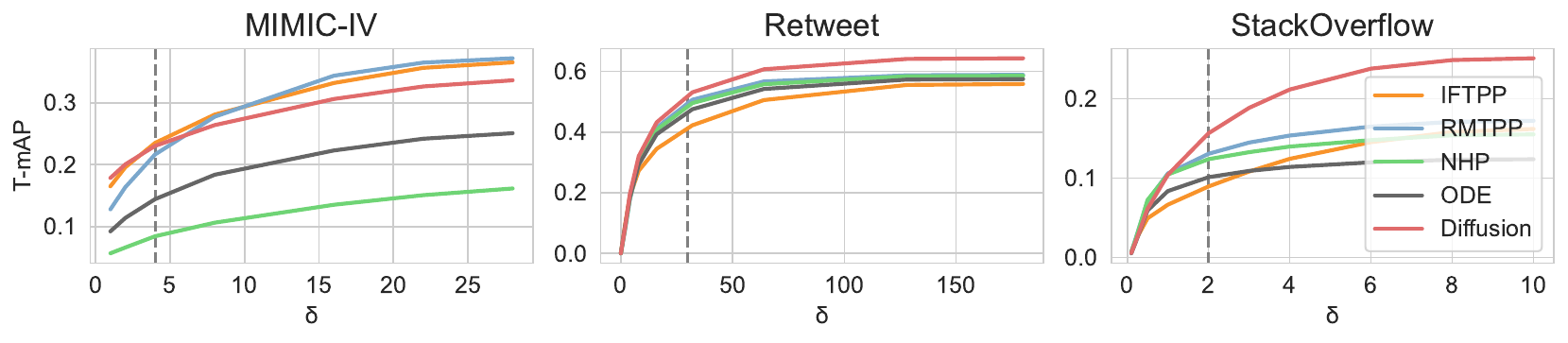}
  \caption{T-mAP dependency on the $\delta$ parameter. The dashed line indicates the selected value. The results for all 5 datasets are presented in \ref{app:tmap-hopt-full}.}
  \label{fig:tmap-hyper}
\end{figure*}

T-mAP has two hyperparameters: the maximum allowed time delta $\delta$ and evaluation horizon $H$. We set $\delta$ twice the cost of the OTD because when the model predicts an incorrect label, OTD removes the prediction and adds the ground truth event with the total cost equal to $2C$. The horizon $H$ must be larger than $\delta$ to evaluate timestamp quality adequately. Therefore, depending on the dataset, we select $H$ to be approximately 3-7 times larger than $\delta$, ensuring the horizon captures an average of 6-15 events. The empirical study of T-mAP hyperparameters is presented in Fig.~\ref{fig:tmap-hyper}. As shown, the chosen $\delta$ parameter is positioned to the right of the initial slope of the parameter-quality curve, ensuring an optimal balance between time granularity and task difficulty.

\subsubsection{Additional T-mAP Studies}
We analyze simple synthetic datasets to further investigate the properties of the proposed T-mAP metric. In \ref{app:irregular}, we demonstrate that T-mAP more effectively distinguishes trivial baselines from statistical methods in scenarios characterized by highly irregular time steps. Additionally, the results in \ref{app:long-tail} show that, unlike OTD, T-mAP captures model performance on rare (long-tail) classes.

\section{HoTPP Benchmark}
\label{sec:benchmark}
The HoTPP benchmark integrates data preprocessing, training, and evaluation in a single toolbox. Unlike previous benchmarks, HoTPP introduces the novel T-mAP metric for long-horizon prediction. HoTPP also differs from prior MTPP benchmarks by including simple rule-based baselines and long-horizon prediction methods that simultaneously predict multiple future events, such as HYPRO~\cite{xue2022hypro} and Diffusion~\cite{zhou2025diffusiontpp}. The HoTPP benchmark is designed to focus on simplicity, extensibility, evaluation stability, reproducibility, and computational efficiency.

{\bf Simplicity and Extensibility.} The benchmark code is organized into a clear structure, separating the core library, dataset-specific scripts, and configuration files. New methods can be integrated at various levels, including the configuration file, model architecture, loss function, metric, and training module. Core components are reusable through the Hydra configuration library \cite{Yadan2019Hydra}. For example, HoTPP can apply the HYPRO rescoring method to backbone models such as IFTPP, RMTPP, NHP, or ODE. Furthermore, each method's backbone architecture can be easily switched between RNNs and Transformers. Methods such as IFTPP, RMTPP, and Diffusion are compatible with most HuggingFace models \cite{huggingface}.

{\bf Evaluation Stability.} In the MTPP domain, many datasets contain only a few thousand sequences for training and evaluation. For example, the StackOverflow test set includes 401 sequences, Retweet contains 1956, and Amazon has 1851 sequences for testing. This possesses a unique challenge in the long-horizon forecasting task. Previous long-horizon evaluation pipelines \cite{xue2022hypro,xue2023easytpp} made predictions only at the end of each sequence, resulting in a limited number of measurements and a low evaluation stability. To address this limitation, we evaluate long-horizon predictions at multiple intermediate points, decreasing the variance of evaluation metrics.


{\bf Reproducibility.} The HoTPP benchmark ensures reproducibility at multiple levels. First, we use the PytorchLightning library \cite{Falcon_PyTorch_Lightning_2019} for training with a specified random seed and report multi-seed evaluation results. Second, data preprocessing is carefully designed to ensure datasets are constructed reproducibly. Finally, we specify the environment in a Dockerfile.

{\bf Computational Efficiency.} Some methods are particularly slow, especially during autoregressive inference on large datasets. Straightforward evaluation can take several hours on a single Nvidia V100 GPU. We optimized the training and inference pipelines in two ways to accelerate computation. First, we implemented an efficient RNN that reuses computations during parallel autoregressive inference from multiple starting positions. Second, we developed highly optimized versions of the thinning algorithm used for sampling in NHP and continuous-time neural architectures (NHP and ODE)~\cite{rizoiu2017hawkes-tutorial}, achieving up to a 4x performance improvement compared to the official implementations. These optimizations allowed us to conduct the first large-scale evaluations of algorithms such as NHP, AttNHP, ODE, and HYPRO on datasets like Transactions and MIMIC-IV. Additionally, we provide the first open-source CUDA implementation of the batched linear assignment solver, significantly enhancing the applicability of the proposed T-mAP metric.

A detailed description of the benchmark can be found in \ref{app:benchmark}, with further details on HoTPP's performance optimizations outlined in \ref{app:performance-improvements}.

\begin{table*}[t]
\caption{Datasets statistics}
\begin{adjustbox}{width=\textwidth}
\begin{tabular}{l|c|c|c|c|c|cc|cc}
\toprule
\multirow{2}{*}{Dataset} & \multirow{2}{*}{\# Sequences} & \multirow{2}{*}{\# Events} & \multirow{2}{*}{Mean Length} & Mean Horizon & \multirow{2}{*}{\# Classes} & \multicolumn{2}{c|}{OTD} & \multicolumn{2}{c}{T-mAP} \\
&&&&Length & & Steps & Cost & $\delta$ & Horizon \\
\midrule
Transactions & 50k & 43.7M & 875 & 9.0 & 203 & 5 & 1 & 2 & 7 \\
MIMIC-IV & 120k & 2.4M & 19.7 & 6.6 & 34 & 5 & 2 & 4 & 28 \\
Retweet & 23k & 1.3M & 56.4 & 14.7 & 3 & 10 & 15 & 30 & 180 \\
Amazon & 9k & 403K & 43.6 & 14.8 & 16 & 5 & 1 & 2 & 10 \\
StackOverflow & 2k & 138K & 64.2 & 12.0 & 22 & 10 & 1 & 2 & 10 \\
\bottomrule
\end{tabular}
\end{adjustbox}
\label{tab:datasets}
\end{table*}

\subsection{Datasets} We for the first time combine datasets from a wide range of domains such as financial transactions, social networks, and medical records into a single evaluation benchmark. Specifically, we provide evaluation results on a transactional dataset \cite{babaev2022coles}, the MIMIC-IV medical dataset \cite{johnson2020mimic}, and social network datasets (Retweet \cite{zhao2015seismic}, StackOverflow \cite{jure2014snap}, and Amazon \cite{jure2014amazon}). These datasets represent diverse underlying processes: social network data is influenced by cascades \cite{zhao2015seismic}, medical records exhibit repetitive patterns and transactional data reflects daily activities, combining regularity with significant uncertainty.

The dataset statistics are presented in Table \ref{tab:datasets}. The Transactions dataset has the longest average sequence length and the largest number of classes, while MIMIC-IV contains the highest number of sequences. Retweet is medium size, whereas Amazon and StackOverflow can be considered small datasets. Further datasets details are reported in \ref{app:benchmark} and \ref{app:domain-analysis}.

\subsection{Methods}
We implement representatives from different groups, listed below.
\begin{itemize}
\item \textbf{Statistical baselines.} \textit{MostPopular} generates K future events with average time steps. The predicted labels approximate distribution of labels in historical data. The \textit{HistoryDistribution} baseline resembles MostPopular, but estimates types densities w.r.t. time features.

\item \textbf{Intensity-free approaches.} We implement the \textit{IFTPP} method, which combines mean absolute error (MAE) of the time step prediction with cross-entropy categorical loss for labels \cite{shchur2019intensityfree,padhi2021tabgpt, mcdermott2024eventstreamgpt}. IFTPP uses GRU reccurrent network as a backbone~\cite{cho2014gru}. We also evaluate IFTPP-T model with a transformer backbone~\cite{radford2019gpt2}.
    
\item \textbf{Intensity-based approaches.} We implement \textit{RMTPP}~\cite{du2016rmtpp} as an example of the TPP approach with a traditional RNN. We add \textit{NHP}~\cite{mei2017nhp}, based on a continuous time LSTM architecture. We also evaluate the \textit{AttNHP} approach that utilizes a continuous time transformer model \cite{yang2022anhp} and \textit{ODE}~\cite{rubanova2019ode}, which applies ordinary differential equations for temporal dynamics modeling.
    
    
\item \textbf{Reranking.} We implement \textit{HYPRO}~\cite{xue2022hypro}, which generates multiple hypotheses with RMTPP and selects the best sequence using a contrastive approach.
    
\item \textbf{Diffusion.} We include the \textit{Diffusion} approach \cite{zhou2025diffusiontpp}, which extends previous research in language modeling \cite{li2022diffusionlm}. Diffusion model iteratively refines predicted sequence, simultaneously generating all events within the specified horizon. Notably, HoTPP provides the first public implementation of this approach.
\end{itemize}

We also explored ContiFormer~\cite{chen2023contiformer} and NJDTPP~\cite{zhang2024njdtpp}, but excluded them from our final comparison for the reasons detailed in \ref{app:benchmark}.
Additional details on implemented methods are provided in \ref{app:background-tpp}, with training specifics outlined in \ref{app:training} and hyperparameters listed in \ref{app:hyperparameters}.

\begin{table*}[p]
\caption{Evaluation results. The best result is shown in bold. The mean and standard deviation of each metric computed during five runs with different random seeds are reported.}
\begin{adjustbox}{width=\textwidth}
\begin{tabular}{l|l|c|ccc|cc}
\toprule
& \multirow{2}{*}{Method} & \multirow{2}{*}{\thead{Mean \\ length}} & \multicolumn{3}{c|}{Next-event} & \multicolumn{2}{c}{Long-horizon} \\
\cline{4-8}
& & & Acc (\%) & mAP (\%) & MAE & \thead{OTD Val / Test} & \thead{T-mAP Val / Test (\%)} \\
\midrule

\multirow{10}{*}{\bf \parbox[t]{2mm}{{\rotatebox[origin=c]{90}{\large Transactions}}}}

& MostPopular & 5.0 & 32.69{\tiny $\pm$ 0.00 } & 1.09{\tiny $\pm$ 0.00 } & 0.649{\tiny $\pm$ 0.000 } & 7.03{\tiny $\pm$ 0.00 } / 7.05{\tiny $\pm$ 0.00 } & 2.86{\tiny $\pm$ 0.00 } / 3.10{\tiny $\pm$ 0.00 } \\
& HistoryDensity & 154.2 & 32.78{\tiny $\pm$ 0.00 } & 0.85{\tiny $\pm$ 0.00 } & 1.063{\tiny $\pm$ 0.000 } & 7.42{\tiny $\pm$ 0.00 } / 7.44{\tiny $\pm$ 0.00 } & 7.92{\tiny $\pm$ 0.00 } / 7.51{\tiny $\pm$ 0.00 } \\
& IFTPP & 13.3 & 38.27{\tiny $\pm$ 0.03 } & {\bf 4.65{\tiny $\pm$ 0.03 }} & {\bf 0.637{\tiny $\pm$ 0.001 }} & 6.79{\tiny $\pm$ 0.01 } / 6.81{\tiny $\pm$ 0.01 } & 6.20{\tiny $\pm$ 0.10 } / 6.08{\tiny $\pm$ 0.14 } \\
& IFTPP-T & 12.8 & 37.68{\tiny $\pm$ 0.03 } & 4.30{\tiny $\pm$ 0.03 } & 0.640{\tiny $\pm$ 0.001 } & 6.84{\tiny $\pm$ 0.02 } / 6.85{\tiny $\pm$ 0.02 } & 5.80{\tiny $\pm$ 0.27 } / 5.55{\tiny $\pm$ 0.19 } \\
& RMTPP & 8.7 & {\bf 38.28{\tiny $\pm$ 0.04 }} & 4.64{\tiny $\pm$ 0.02 } & 0.693{\tiny $\pm$ 0.005 } & {\bf 6.68{\tiny $\pm$ 0.01 }} / {\bf 6.70{\tiny $\pm$ 0.01 }} & 7.41{\tiny $\pm$ 0.35 } / 7.12{\tiny $\pm$ 0.10 } \\
& NHP & 9.4 & 35.47{\tiny $\pm$ 0.12 } & 3.36{\tiny $\pm$ 0.02 } & 0.693{\tiny $\pm$ 0.005 } & 6.99{\tiny $\pm$ 0.03 } / 6.99{\tiny $\pm$ 0.03 } & 5.66{\tiny $\pm$ 0.17 } / 5.48{\tiny $\pm$ 0.12 } \\
& AttNHP & 6.8 & 30.68{\tiny $\pm$ x.xx } & 1.09{\tiny $\pm$ x.xx } & 0.749{\tiny $\pm$ x.xxx } & 7.47{\tiny $\pm$ x.xx } / 7.48{\tiny $\pm$ x.xx } & 1.50{\tiny $\pm$ x.xx } / 1.49{\tiny $\pm$ x.xx } \\
& ODE & 9.1 & 35.60{\tiny $\pm$ 0.06 } & 3.34{\tiny $\pm$ 0.06 } & 0.695{\tiny $\pm$ 0.002 } & 6.96{\tiny $\pm$ 0.01 } / 6.97{\tiny $\pm$ 0.01 } & 5.53{\tiny $\pm$ 0.08 } / 5.52{\tiny $\pm$ 0.13 } \\
& HYPRO & 7.3 & 22.50{\tiny $\pm$ x.xx } & 4.62{\tiny $\pm$ x.xx } & 0.694{\tiny $\pm$ x.xxx } & 7.62{\tiny $\pm$ x.xx } / 7.62{\tiny $\pm$ x.xx } & {\bf 8.20{\tiny $\pm$ x.xx }} / {\bf 7.54{\tiny $\pm$ x.xx }} \\
& Diffusion & 10.4 & 36.05{\tiny $\pm$ 0.07 } & 3.66{\tiny $\pm$ 0.03 } & 0.650{\tiny $\pm$ 0.002 } & 6.86{\tiny $\pm$ 0.01 } / 6.88{\tiny $\pm$ 0.00 } & 5.95{\tiny $\pm$ 0.05 } / 6.04{\tiny $\pm$ 0.08 } \\

\midrule

\multirow{10}{*}{\bf \parbox[t]{2mm}{{\rotatebox[origin=c]{90}{\large MIMIC-IV}}}}

& MostPopular & 2.7 & 4.80{\tiny $\pm$ 0.00 } & 2.75{\tiny $\pm$ 0.00 } & 14.52{\tiny $\pm$ 0.00 } & 19.75{\tiny $\pm$ 0.00 } / 19.75{\tiny $\pm$ 0.00 } & 1.63{\tiny $\pm$ 0.00 } / 1.56{\tiny $\pm$ 0.00 } \\
& HistoryDensity & 160.0 & 0.74{\tiny $\pm$ 0.00 } & 2.63{\tiny $\pm$ 0.00 } & 3.23{\tiny $\pm$ 0.00 } & 18.94{\tiny $\pm$ 0.00 } / 18.92{\tiny $\pm$ 0.00 } & 4.25{\tiny $\pm$ 0.00 } / 4.19{\tiny $\pm$ 0.00 } \\
& IFTPP & 12.4 & 58.52{\tiny $\pm$ 0.06 } & 46.04{\tiny $\pm$ 0.26 } & 3.00{\tiny $\pm$ 0.02 } & 11.53{\tiny $\pm$ 0.03 } / 11.55{\tiny $\pm$ 0.03 } & 22.56{\tiny $\pm$ 0.13 } / 22.31{\tiny $\pm$ 0.15 } \\
& IFTPP-T & 11.7 & {\bf 58.57{\tiny $\pm$ 0.04 }} & {\bf 47.96{\tiny $\pm$ 1.51 }} & 3.03{\tiny $\pm$ 0.01 } & {\bf 11.46{\tiny $\pm$ 0.02 }} / {\bf 11.48{\tiny $\pm$ 0.02 }} & {\bf 24.07{\tiny $\pm$ 0.31 }} / {\bf 23.68{\tiny $\pm$ 0.29 }} \\
& RMTPP & 9.8 & 58.24{\tiny $\pm$ 0.14 } & 45.62{\tiny $\pm$ 1.28 } & 3.93{\tiny $\pm$ 0.03 } & 13.69{\tiny $\pm$ 0.05 } / 13.76{\tiny $\pm$ 0.06 } & 22.00{\tiny $\pm$ 0.28 } / 21.62{\tiny $\pm$ 0.29 } \\
& NHP & 3.0 & 23.19{\tiny $\pm$ 0.61 } & 11.67{\tiny $\pm$ 0.39 } & 4.71{\tiny $\pm$ 0.31 } & 17.92{\tiny $\pm$ 0.34 } / 17.92{\tiny $\pm$ 0.33 } & 9.76{\tiny $\pm$ 0.41 } / 9.80{\tiny $\pm$ 0.38 } \\
& AttNHP & 6.2 & 44.99{\tiny $\pm$ 1.11 } & 31.88{\tiny $\pm$ 0.64 } & 2.95{\tiny $\pm$ 0.20 } & 14.30{\tiny $\pm$ 0.16 } / 14.31{\tiny $\pm$ 0.16 } & 19.98{\tiny $\pm$ 0.79 } / 19.74{\tiny $\pm$ 0.71 } \\
& ODE & 11.6 & 44.16{\tiny $\pm$ 0.76 } & 27.48{\tiny $\pm$ 1.04 } & {\bf 2.91{\tiny $\pm$ 0.01 }} & 14.36{\tiny $\pm$ 0.20 } / 14.39{\tiny $\pm$ 0.20 } & 16.37{\tiny $\pm$ 0.75 } / 16.19{\tiny $\pm$ 0.74 } \\
& HYPRO & 7.4 & 45.69{\tiny $\pm$ x.xx } & 45.54{\tiny $\pm$ x.xx } & 3.98{\tiny $\pm$ x.xx } & 14.68{\tiny $\pm$ x.xx } / 14.75{\tiny $\pm$ x.xx } & 16.41{\tiny $\pm$ x.xx } / 16.13{\tiny $\pm$ x.xx } \\
& Diffusion & 15.4 & 46.22{\tiny $\pm$ 5.31 } & 33.67{\tiny $\pm$ 3.66 } & 2.99{\tiny $\pm$ 0.01 } & 13.25{\tiny $\pm$ 0.15 } / 13.28{\tiny $\pm$ 0.14 } & 22.90{\tiny $\pm$ 0.10 } / 22.82{\tiny $\pm$ 0.12 } \\

\midrule

\multirow{10}{*}{\bf \parbox[t]{2mm}{{\rotatebox[origin=c]{90}{\large Retweet}}}}

& MostPopular & 9.9 & 58.50{\tiny $\pm$ 0.00 } & 39.85{\tiny $\pm$ 0.00 } & 18.82{\tiny $\pm$ 0.00 } & {\bf 141.7{\tiny $\pm$ 0.0 }} / {\bf 139.1{\tiny $\pm$ 0.0 }} & 29.72{\tiny $\pm$ 0.00 } / 30.02{\tiny $\pm$ 0.00 } \\
& HistoryDensity & 14.9 & 58.62{\tiny $\pm$ 0.00 } & 39.86{\tiny $\pm$ 0.00 } & 20.68{\tiny $\pm$ 0.00 } & 178.5{\tiny $\pm$ 0.0 } / 177.0{\tiny $\pm$ 0.0 } & 44.48{\tiny $\pm$ 0.00 } / 44.60{\tiny $\pm$ 0.00 } \\
& IFTPP & 23.1 & 59.75{\tiny $\pm$ 0.25 } & 46.23{\tiny $\pm$ 0.31 } & {\bf 18.27{\tiny $\pm$ 0.05 }} & 165.8{\tiny $\pm$ 0.2 } / 164.8{\tiny $\pm$ 0.3 } & 43.93{\tiny $\pm$ 0.69 } / 40.59{\tiny $\pm$ 0.84 } \\
& IFTPP-T & 23.2 & 59.88{\tiny $\pm$ 0.06 } & 46.07{\tiny $\pm$ 0.11 } & 18.47{\tiny $\pm$ 0.06 } & 167.4{\tiny $\pm$ 1.3 } / 166.1{\tiny $\pm$ 1.1 } & 43.73{\tiny $\pm$ 1.88 } / 40.05{\tiny $\pm$ 1.73 } \\
& RMTPP & 16.6 & 60.01{\tiny $\pm$ 0.06 } & 46.81{\tiny $\pm$ 0.03 } & 18.60{\tiny $\pm$ 0.04 } & 169.2{\tiny $\pm$ 1.0 } / 168.0{\tiny $\pm$ 1.0 } & 49.03{\tiny $\pm$ 0.15 } / 45.60{\tiny $\pm$ 0.28 } \\
& NHP & 18.1 & 60.05{\tiny $\pm$ 0.08 } & {\bf 46.84{\tiny $\pm$ 0.13 }} & 18.39{\tiny $\pm$ 0.12 } & 167.0{\tiny $\pm$ 1.7 } / 165.7{\tiny $\pm$ 1.7 } & 49.25{\tiny $\pm$ 0.37 } / 45.64{\tiny $\pm$ 0.26 } \\
& AttNHP & 29.8 & {\bf 60.06{\tiny $\pm$ 0.03 }} & 46.66{\tiny $\pm$ 0.02 } & 18.32{\tiny $\pm$ 0.02 } & 174.7{\tiny $\pm$ 0.7 } / 172.9{\tiny $\pm$ 0.7 } & 26.74{\tiny $\pm$ 1.16 } / 24.24{\tiny $\pm$ 0.94 } \\
& ODE & 17.7 & 59.93{\tiny $\pm$ 0.06 } & 46.36{\tiny $\pm$ 0.30 } & 18.53{\tiny $\pm$ 0.09 } & 168.2{\tiny $\pm$ 1.1 } / 166.7{\tiny $\pm$ 1.0 } & 48.95{\tiny $\pm$ 0.30 } / 45.45{\tiny $\pm$ 0.38 } \\
& HYPRO & 16.2 & 50.84{\tiny $\pm$ x.xx } & {\bf 46.84{\tiny $\pm$ x.xx }} & 18.61{\tiny $\pm$ x.xx } & 161.8{\tiny $\pm$ x.x } / 160.4{\tiny $\pm$ x.x } & 50.58{\tiny $\pm$ x.xx } / 46.87{\tiny $\pm$ x.xx } \\
& Diffusion & 14.7 & 57.72{\tiny $\pm$ 0.77 } & 42.96{\tiny $\pm$ 0.67 } & 18.60{\tiny $\pm$ 0.06 } & 159.0{\tiny $\pm$ 1.4 } / 158.0{\tiny $\pm$ 1.1 } & {\bf 54.24{\tiny $\pm$ 0.89 }} / {\bf 52.24{\tiny $\pm$ 0.68 }} \\

\midrule

\multirow{10}{*}{\bf \parbox[t]{2mm}{{\rotatebox[origin=c]{90}{\large Amazon}}}}

& MostPopular & 5.0 & 33.46{\tiny $\pm$ 0.00 } & 9.58{\tiny $\pm$ 0.00 } & 0.304{\tiny $\pm$ 0.000 } & {\bf 6.33{\tiny $\pm$ 0.00 }} / {\bf 6.30{\tiny $\pm$ 0.00 }} & 10.10{\tiny $\pm$ 0.00 } / 9.72{\tiny $\pm$ 0.00 } \\
& HistoryDensity & 71.8 & 32.68{\tiny $\pm$ 0.00 } & 9.63{\tiny $\pm$ 0.00 } & 0.325{\tiny $\pm$ 0.000 } & 7.29{\tiny $\pm$ 0.00 } / 7.27{\tiny $\pm$ 0.00 } & 28.25{\tiny $\pm$ 0.00 } / 27.53{\tiny $\pm$ 0.00 } \\
& IFTPP & 12.8 & 35.54{\tiny $\pm$ 0.08 } & 16.72{\tiny $\pm$ 0.04 } & 0.250{\tiny $\pm$ 0.001 } & 6.70{\tiny $\pm$ 0.01 } / 6.64{\tiny $\pm$ 0.01 } & 23.05{\tiny $\pm$ 0.48 } / 23.79{\tiny $\pm$ 0.39 } \\
& IFTPP-T & 13.1 & 35.27{\tiny $\pm$ 0.04 } & 16.46{\tiny $\pm$ 0.07 } & {\bf 0.245{\tiny $\pm$ 0.001 }} & 6.64{\tiny $\pm$ 0.02 } / 6.59{\tiny $\pm$ 0.02 } & 22.81{\tiny $\pm$ 0.10 } / 23.42{\tiny $\pm$ 0.13 } \\
& RMTPP & 16.4 & {\bf 35.85{\tiny $\pm$ 0.03 }} & {\bf 17.21{\tiny $\pm$ 0.04 }} & 0.292{\tiny $\pm$ 0.002 } & 6.67{\tiny $\pm$ 0.03 } / 6.62{\tiny $\pm$ 0.03 } & 19.65{\tiny $\pm$ 0.23 } / 19.99{\tiny $\pm$ 0.18 } \\
& NHP & 9.9 & 16.58{\tiny $\pm$ 3.26 } & 11.19{\tiny $\pm$ 0.18 } & 0.438{\tiny $\pm$ 0.009 } & 8.54{\tiny $\pm$ 0.28 } / 8.52{\tiny $\pm$ 0.30 } & 27.19{\tiny $\pm$ 0.35 } / 26.95{\tiny $\pm$ 0.37 } \\
& AttNHP & 12.4 & 22.21{\tiny $\pm$ 5.64 } & 8.22{\tiny $\pm$ 0.39 } & 0.526{\tiny $\pm$ 0.075 } & 7.63{\tiny $\pm$ 0.12 } / 7.61{\tiny $\pm$ 0.14 } & 19.29{\tiny $\pm$ 1.45 } / 18.59{\tiny $\pm$ 1.52 } \\
& ODE & 9.4 & 10.44{\tiny $\pm$ 2.84 } & 10.28{\tiny $\pm$ 0.57 } & 0.473{\tiny $\pm$ 0.031 } & 9.23{\tiny $\pm$ 0.17 } / 9.23{\tiny $\pm$ 0.18 } & 23.73{\tiny $\pm$ 0.36 } / 23.13{\tiny $\pm$ 0.46 } \\
& HYPRO & 17.8 & 21.47{\tiny $\pm$ x.xx } & 17.14{\tiny $\pm$ x.xx } & 0.288{\tiny $\pm$ x.xxx } & 6.95{\tiny $\pm$ x.xx } / 6.95{\tiny $\pm$ x.xx } & 20.47{\tiny $\pm$ x.xx } / 20.28{\tiny $\pm$ x.xx } \\
& Diffusion & 13.3 & 30.57{\tiny $\pm$ 1.28 } & 13.36{\tiny $\pm$ 0.49 } & 0.247{\tiny $\pm$ 0.000 } & 6.56{\tiny $\pm$ 0.03 } / 6.52{\tiny $\pm$ 0.04 } & {\bf 30.08{\tiny $\pm$ 0.25 }} / {\bf 30.29{\tiny $\pm$ 0.32 }} \\

\midrule

\multirow{10}{*}{\bf \parbox[t]{2mm}{{\rotatebox[origin=c]{90}{\large StackOverflow}}}}

& MostPopular & 9.8 & 42.90{\tiny $\pm$ 0.00 } & 5.45{\tiny $\pm$ 0.00 } & {\bf 0.602{\tiny $\pm$ 0.000 }} & 13.12{\tiny $\pm$ 0.00 } / 13.15{\tiny $\pm$ 0.00 } & 8.97{\tiny $\pm$ 0.00 } / 8.16{\tiny $\pm$ 0.00 } \\
& HistoryDensity & 74.0 & 42.51{\tiny $\pm$ 0.00 } & 5.61{\tiny $\pm$ 0.00 } & 0.669{\tiny $\pm$ 0.000 } & 17.30{\tiny $\pm$ 0.00 } / 17.29{\tiny $\pm$ 0.00 } & {\bf 16.23{\tiny $\pm$ 0.00 }} / 14.90{\tiny $\pm$ 0.00 } \\
& IFTPP & 24.1 & 45.31{\tiny $\pm$ 0.10 } & 11.59{\tiny $\pm$ 0.70 } & 0.646{\tiny $\pm$ 0.003 } & 13.56{\tiny $\pm$ 0.19 } / 13.66{\tiny $\pm$ 0.19 } & 9.68{\tiny $\pm$ 0.55 } / 8.87{\tiny $\pm$ 0.59 } \\
& IFTPP-T & 23.0 & {\bf 45.40{\tiny $\pm$ 0.04 }} & {\bf 12.78{\tiny $\pm$ 0.29 }} & 0.639{\tiny $\pm$ 0.001 } & 13.51{\tiny $\pm$ 0.07 } / 13.61{\tiny $\pm$ 0.05 } & 9.62{\tiny $\pm$ 0.15 } / 8.81{\tiny $\pm$ 0.20 } \\
& RMTPP & 12.7 & 45.30{\tiny $\pm$ 0.07 } & 12.61{\tiny $\pm$ 0.40 } & 0.717{\tiny $\pm$ 0.006 } & 13.00{\tiny $\pm$ 0.03 } / 13.25{\tiny $\pm$ 0.04 } & 13.81{\tiny $\pm$ 0.30 } / 13.35{\tiny $\pm$ 0.28 } \\
& NHP & 11.9 & 44.12{\tiny $\pm$ 0.79 } & 9.90{\tiny $\pm$ 1.16 } & 0.735{\tiny $\pm$ 0.021 } & 13.27{\tiny $\pm$ 0.31 } / 13.46{\tiny $\pm$ 0.27 } & 13.12{\tiny $\pm$ 0.26 } / 12.41{\tiny $\pm$ 0.31 } \\
& AttNHP & 15.6 & 45.17{\tiny $\pm$ 0.16 } & 12.56{\tiny $\pm$ 0.15 } & 0.703{\tiny $\pm$ 0.000 } & 13.10{\tiny $\pm$ 0.03 } / 13.31{\tiny $\pm$ 0.01 } & 11.95{\tiny $\pm$ 0.38 } / 11.15{\tiny $\pm$ 0.29 } \\
& ODE & 13.3 & 44.19{\tiny $\pm$ 0.13 } & 9.08{\tiny $\pm$ 0.27 } & 0.728{\tiny $\pm$ 0.007 } & 13.12{\tiny $\pm$ 0.05 } / 13.35{\tiny $\pm$ 0.05 } & 11.48{\tiny $\pm$ 0.29 } / 10.75{\tiny $\pm$ 0.21 } \\
& HYPRO & 11.8 & 30.54{\tiny $\pm$ 0.70 } & 12.61{\tiny $\pm$ 0.40 } & 0.718{\tiny $\pm$ 0.006 } & 14.00{\tiny $\pm$ 0.19 } / 14.12{\tiny $\pm$ 0.17 } & 15.25{\tiny $\pm$ 0.28 } / 14.84{\tiny $\pm$ 0.23 } \\
& Diffusion & 14.4 & 37.03{\tiny $\pm$ 1.24 } & 7.48{\tiny $\pm$ 0.16 } & 0.668{\tiny $\pm$ 0.006 } & {\bf 12.87{\tiny $\pm$ 0.17 }} / {\bf 13.01{\tiny $\pm$ 0.16 }} & 15.80{\tiny $\pm$ 0.85 } / {\bf 15.07{\tiny $\pm$ 0.74 }} \\

\bottomrule
\end{tabular}
\end{adjustbox}
\label{tab:results}
\end{table*}

\section{Experiments}
\label{sec:experiments}

The main evaluation results are presented in Table~\ref{tab:results}. In the following sections, we will discuss these results from various perspectives and offer additional analysis of the methods' behavior. 


\subsection{Long-horizon Prediction Quality}
\label{exp:long-horizon}
According to Table \ref{tab:results}, autoregressive approaches such as IFTPP, RMTPP, and NHP excel at next-event prediction (MAE and accuracy), while horizon prediction methods like HYPRO and Diffusion perform better in terms of OTD and T-mAP. These findings indicate that next-event and horizon prediction are distinct tasks requiring specialized models, and none of the existing approaches achieve top performance in both domains.

Our statistical baselines rank first in 3 out of 10 long-horizon prediction tasks, and in three additional tasks they perform on par with the top methods without a statistically significant difference. These findings highlight the need for greater focus on long-horizon forecasting and point to opportunities for future improvements.

Intensity-based approaches such as RMTPP and NHP generally outperform the autoregressive, intensity-free strategy. Meanwhile, Diffusion relies on intensity-free losses, suggesting that integrating intensity-based loss functions into Diffusion could potentially enhance long-horizon prediction.

Notably, Transformer-based methods rarely offer advantages over RNN architectures. For instance, IFTPP-T significantly surpasses the RNN-based IFTPP only on MIMIC-IV, while being outperformed in most other scenarios. Consequently, the potential of Transformer architectures in the MTPP domain remains insufficiently explored.

\subsection{Metrics Comparison}
\label{exp:metric}
Table \ref{tab:results} shows that the OTD metric can yield low values even for simple rule-based baselines. For example, the MostPopular baseline achieves the lowest OTD scores on the Retweet and Amazon datasets. In contrast, T-mAP more clearly distinguishes between rule-based baselines and deep learning models by measuring predicted labels probabilities rather than hard labels.


The difference between OTD and T-mAP also emerges when comparing autoregressive methods that use the NHP loss (NHP, ODE) with those that jointly model timestamp and label for the next event (IFTPP, RMTPP). For instance, NHP performs poorly on OTD yet ranks among the top autoregressive approaches by T-mAP. This discrepancy arises from how the two metrics balance event order and time prediction quality. OTD considers only the first 5--10 events, so the model must accurately predict these initial events. Because IFTPP and RMTPP jointly predict the next event’s time and label, they generally preserve ordering more effectively. In contrast, NHP, AttNHP, and ODE predict times separately for each event type, causing random ordering when timestamps are close and thereby lowering the OTD score. We conclude that T-mAP is less sensitive to event order as long as the predicted times are accurate.

The HYPRO and Diffusion methods, specifically designed for long-horizon prediction, achieve high T-mAP scores but often underperform autoregressive approaches in terms of OTD. This highlights T-mAP's ability to reward models that produce realistic sequences. We also observe that methods with high T-mAP scores frequently have low next-event prediction accuracy. For example, on the Retweet, Amazon, and StackOverflow datasets, high long-horizon performance often coincides with low next-event prediction scores. This is especially true for HYPRO and Diffusion, whose loss functions heavily emphasize long-horizon quality. Meanwhile, the strong correlation of next-event prediction quality and long-horizon performance on MIMIC-IV can be attributed to the large number of events with identical timestamps, which makes precise event ordering crucial and favors autoregressive approaches like IFTPP.



\subsection{Mode Collapse}
\label{sec:collapse}

\begin{figure*}[t]
\centering
  \includegraphics[width=\textwidth]{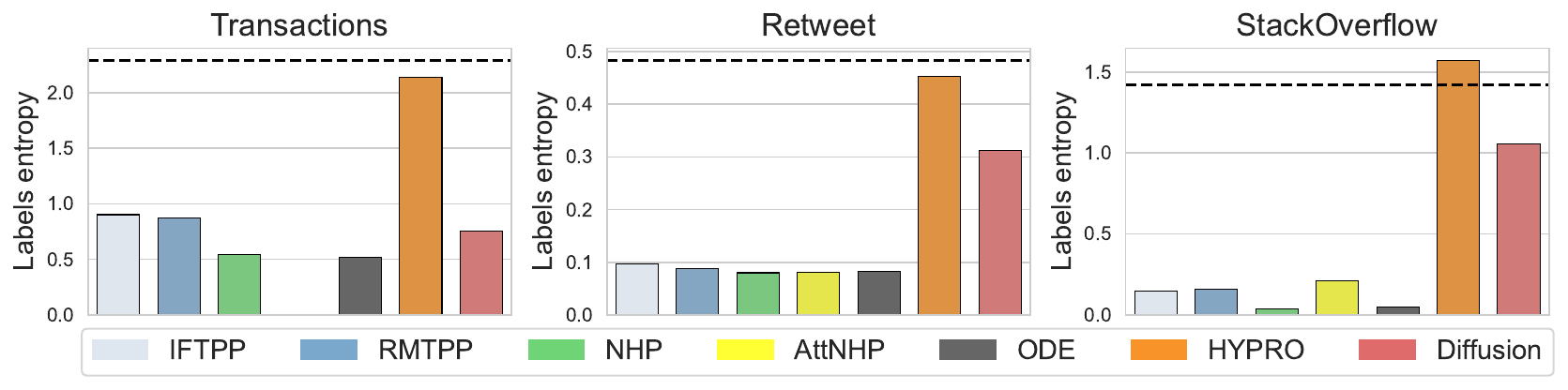}
  \vspace{-0.2in}
  \caption{Comparison of predicted label entropies for different methods, with the dashed horizontal line indicating the ground truth entropy. The results for all 5 datasets are presented in \ref{app:qualitative-diversity}.}
  \label{fig:entropies}
\end{figure*}

We found that most models tend to predict a single label or repetitive patterns in long-horizon tasks (see examples from \ref{app:qualitative-diversity}).
To quantify the effect of mode collapse, we measured the entropies of predicted labels, as illustratedin Figure~\ref{fig:entropies}. Autoregressive methods generally exhibit low prediction diversity across all datasets except MIMIC-IV, where strong dependencies enable the model to learn typical sequences. Meanwhile, HYPRO, which samples from RMTPP with an increased temperature, substantially enhances label diversity. Diffusion similarly boosts diversity except on the Transactions dataset. Overall, most methods, except HYPRO, experience mode collapse in long-horizon prediction, although horizon-based approaches partly mitigate this issue. Sampling with an increased temperature can potentially solve this issue by offering a trade-off between predictions quality and diversity.

\section{Conclusion}

In this paper, we proposed the HoTPP benchmark to assess the quality of long-horizon events forecasting. HoTPP provides optimized implementations of the state-of-the-art methods ranging from statistical baselines to ODE and Diffusion (Sec.\ref{sec:benchmark}), along with diverse metrics for both next-event and long-horizon prediction (Sec.\ref{sec:metric}). Our extensive evaluations using diverse datasets and various predictive models reveal a critical insight: high accuracy in next-event prediction typically does not correlate with superior performance in horizon prediction (Sec.~\ref{exp:long-horizon}). Moreover, most existing methods suffer from mode collapse when forecasting longer sequences (Sec.~\ref{sec:collapse}). These findings underscore the need for developing specialized models. By emphasizing long-term predictive capabilities, HoTPP aims to stimulate the development of more robust and reliable event-sequence models. This advance has the potential to substantially enhance practical applications in various industries, driving both innovation and more informed decision-making.

\section{Limitations and Future Work}
The proposed benchmark has several shortcomings. The training process for some methods is relatively slow. For example, a continuous time LSTM from the NHP method lacks an effective open-source GPU implementation, which could be developed in the future. Autoregression can also be optimized using specialized GPU implementations, which would significantly impact HYPRO training. Additionally, the list of implemented methods can be extended, for example, by incorporating multi-task models \cite{deshpande2021dualtpp}.

Our benchmark encourages future research to develop methods capable of solving next-event and long-horizon prediction tasks simultaneously, as discussed in Section~\ref{exp:long-horizon}. One promising direction involves adapting concepts from our statistical baselines for deep models, thereby surpassing baseline performance. A key question in this effort is the correct alignment between predicted and ground truth events during training, as discussed in the context of T-mAP (Section~\ref{sec:metric}). Future work could also target the reduction of mode collapse and the enhancement of predictions diversity, as highlighted in Section~\ref{sec:collapse}. 

On the other hand, the T-mAP metric can be applied in domains beyond event sequences, such as action recognition \cite{su2017orderwasserstein}. T-mAP can potentially offer better timestamp evaluation in these domains than evaluating indices with OPW. T-mAP can also provide a more natural hyperparameter selection regarding the modeling horizon and maximum allowed time error. Our theoretical justification of T-mAP can potentially be adapted for mAP estimation algorithms in computer vision.


\printbibliography

\appendix

\theoremstyle{plain}
\newtheorem{manualtheoreminner}{Theorem}
\newenvironment{manualtheorem}[1]{%
  \renewcommand\themanualtheoreminner{#1}%
  \manualtheoreminner
}{\endmanualtheoreminner}

\theoremstyle{plain}
\newtheorem{manuallemmainner}{Lemma}
\newenvironment{manuallemma}[1]{%
  \renewcommand\themanuallemmainner{#1}%
  \manuallemmainner
}{\endmanuallemmainner}

\section{Modeling Marked Temporal Point Processes}
\label{app:background-tpp}
{\bf Intensity-based approaches.}
Modeling the probability density function (PDF) $f^*(t_i) = f(t_i | t_1, \dots, t_{i - 1})$ for the next event time is typically challenging, as it requires the additional constraint that its integral equals one. Instead, the non-negative intensity function $\lambda(t_i) \ge 0$ is usually modeled. The following equation gives the relationship between the PDF and the intensity function:
\begin{equation}
    f^*(t_i) = \lambda(t_i)\exp\left(-\int_{t_{i - 1}}^{t_i} \lambda(s)ds\right).
\end{equation}

The derivation is provided in \cite{rizoiu2017hawkes-tutorial}. Different TPPs are characterized by their intensity function $\lambda(t_i)$. In Poisson and non-homogeneous Poisson processes, the intensity function is independent of previous events, meaning event occurrences depend solely on external factors. In contrast, self-exciting processes are characterized by previous events increasing the intensity of future events. A notable example of a self-exciting process is the Hawkes process, in which each event linearly affects the future intensity:
\begin{equation}
    \lambda(t_i) = \lambda_0(t_i) + \sum\limits_{k = 1}^{i - 1} \phi (t_i - t_k),
\end{equation}

where $\lambda_0(t_i) \ge 0$ is the base intensity function, independent of previous events, and $\phi(x) \ge 0$ is the so-called {\it memory kernel} function. Given the intensity function $\lambda(t_i)$, predictions are typically made by sampling or expectation estimation. Sampling is usually performed using the {\it thinning algorithm}, a specific rejection sampling approach. For details on the implementation of sampling, please refer to \cite{rizoiu2017hawkes-tutorial}.


Recent research has focused on modeling complex intensity functions. Various neural network architectures have been adapted to address this problem. These approaches differ in the type of neural network used and the model of the intensity function. Neural architectures range from simple RNNs and Transformers to specially designed continuous-time models like NHP \cite{mei2017nhp} and Neural ODEs \cite{rubanova2019ode}. The intensity function between events can be modeled as a sum of intensities from previous events, as in Hawkes processes, or by directly predicting the inter-event intensity given the context embedding.

{\bf Intensity-free modeling.} Some methods evaluate the next event time distribution without using intensity functions. For example, intensity-free~\cite{shchur2019intensityfree} represents the distribution as a mixture of Gaussians or through normalizing flows. Instead of predicting the distribution directly, other approaches solve a regression problem using MAE or RMSE loss. However, it can be shown that both MAE and RMSE losses are closely related to distribution prediction. Specifically, RMSE is analogous to log-likelihood optimization with a Normal distribution, and MAE is similar to the log-likelihood of a Laplace distribution~\cite{bishop1994mdn}. In our experiments, we evaluated a model trained with MAE loss as an example of an intensity-free method.

{\bf Rescoring with HYPRO.} HYPRO~\cite{xue2022hypro} is an extension applicable to any sequence prediction method capable of sampling. HYPRO takes a pretrained generative model and trains an additional scoring module to select the best sequence from a sample. It is trained with a contrastive loss to distinguish between the generated sequence and the ground truth. HYPRO generates multiple sequences with a background model during inference and selects the best one by maximizing the estimated score. Although HYPRO is intended to improve quality compared to simple sampling, it is unclear whether HYPRO outperforms expectation-based prediction. In our work, we apply HYPRO to the outputs of the RMTPP model.

\textbf{Diffusion.} Recently, diffusion-based approaches have been applied to NLP tasks \cite{li2022diffusionlm}. This idea has been extended to event sequences by adapting best practices from NLP \cite{zhou2025diffusiontpp}. In such methods, events are first mapped to continuous embeddings, where the diffusion process operates, then mapped back to event space. The process begins with embeddings sampled from Gaussian noise and iteratively refines them using historical event embeddings. Finally, the refined embeddings are projected to event times and labels through a softmax-based layer. Training involves a combination of MSE loss in the continuous space, MAE loss for time reconstruction, and cross-entropy loss for label prediction. In HoTPP, we provide the first public implementation of the approach proposed by \cite{zhou2025diffusiontpp}.

\textbf{MostPopular Baseline.} This baseline estimates the average inter-event time and the empirical label distribution from historical data. For next-event prediction, it outputs the average inter-event duration and selects the label with the highest probability. For long-horizon forecasting, it generates events at fixed time intervals and assigns labels such that the predicted distribution aligns with the estimated historical distribution, thus minimizing the discrepancy between predicted and observed label frequencies. Because it assigns hard labels, this method is particularly designed for next-event prediction and minimizing the OTD metric.

\textbf{HistoryDensity Baseline.} Similar to MostPopular, this approach estimates label distributions from historical data. Unlike MostPopular, it focuses on label densities with respect to time rather than normalizing over the total number of events. Consequently, for long-horizon forecasting, it assigns soft probabilities to each predicted event, thereby maximizing the T-mAP score.

\textbf{Methods excluded from comparison.} 
We attempted to incorporate ContiFormer \cite{chen2023contiformer} into our benchmark but encountered multiple challenges. The official implementation provides scripts only for training on a toy spiral dataset, making it impossible to reproduce the method's performance on real-world data. Moreover, training even on this toy example required approximately three hours on an NVIDIA RTX 4060, suggesting that ContiFormer cannot be efficiently scaled to large datasets such as Transactions. In addition, contrary to the claims in the paper, ODE \cite{rubanova2019ode} and continuous-time transformers \cite{yang2022anhp} were already discussed in earlier work. Therefore, we excluded ContiFormer from our benchmark in favor of ODE and AttNHP.

We also attempted to include NJDTPP \cite{zhang2024njdtpp} in our benchmark, but encountered several challenges. Even on StackOverflow, our smallest dataset, the inference time spanned two hours for a single test set prediction. Moreover, after running the official implementation on StackOverflow, we observed accuracy values that were inconsistent with the paper and fell below most baseline results. We also identified data leakage during inference, where future event timestamps were used to predict the subsequent event. Consequently, we chose not to include NJDTPP in our comparisons.



\section{T-mAP Computation and Proofs}
\label{app:tmapcomp}

{\bf Definitions and scope.} In this section, we consider weighted bipartite graphs. The first part consists of {\it predictions}, and the second consists of {\it ground truth} events. We assume that all edges connected to the same prediction have the same weight. A matching is a set of edges $(a_i, b_i)$ between predictions and ground truth events. A matching is termed {\it optimal} if it (1) has the maximum size among all possible matching and (2) has the minimum total weight among all matches of that size. We denote all optimal matchings in the graph $\mathcal{G}$ as $M(\mathcal{G})$.

\begin{manuallemma}{A}
Consider a graph $\mathcal{G}'$, constructed from a graph $\mathcal{G}$ by adding one ground truth vertex with corresponding weighted edges. Then any optimal matching $m' \in M(\mathcal{G}')$ will either have a size greater than the sizes of matchings in $M(\mathcal{G})$ or have a total weight equal to that of matchings in $M(\mathcal{G})$.
\label{theorem:lemma}
\end{manuallemma}
\vspace{-0.3in}
\begin{figure*}[h]
\centering
  \includegraphics[width=\textwidth]{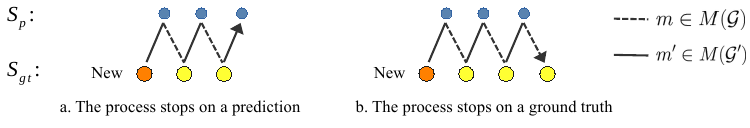}
  \vspace{-0.25in}
  \caption{Illustration of the iterative process in Lemma A.}
  \vspace{-0.2in}
  \label{fig:lemma}
\end{figure*}
\begin{proof}
If the size of $m'$ is greater than the size of matchings in $M(\mathcal{G})$, then the lemma holds. The optimal matching in the extended graph $\mathcal{G}'$ cannot have a size smaller than the optimal matching in $\mathcal{G}$. Now, consider the case when both matchings have the same size. We will show that $m'$ has a total weight equal to that of the matchings in $M(\mathcal{G})$.

Denote $b$ the new ground truth event in graph $\mathcal{G}'$. If $m'$ does not include vertex $b$, then $m'$ is also optimal in $\mathcal{G}$, and the lemma holds. If $m'$ includes vertex $b$, then we can walk through the graph with the following process:
\begin{enumerate}
    \item Start from the new vertex $b$.
    \item If we are at a ground truth event $b_i$ and there is an edge $(a_i, b_i)$ in matching $m'$, move to the vertex $a_i$.
    \item If we are at a prediction $a_i$ and there is an edge $(a_i, b_i)$ in matching $m$, move to the vertex $b_i$.
\end{enumerate}
Example processes are illustrated in Fig.~\ref{fig:lemma}. The vertices in each process do not repeat; otherwise, there would be two edges in either $m$ or $m'$ with the same vertex, which contradicts the definition of a matching.

If the process finishes at a predicted event, then $m'$ has a size greater than the size of $m$, and the lemma holds. If the process finishes at a ground truth event, we can replace a part of matching $m'$ traced with a part of matching $m$. The resulting matching $\hat{m}$ will have a size equal to $m$ and $m'$ and a total weight equal to both matchings. This proves the lemma.
\end{proof}


\begin{manualtheorem}{4.1}
For any threshold $h$, there exists an optimal matching in the graph $\mathcal{G}_h$ such that it is a subset of an optimal matching in the full graph $\mathcal{G}$:
\begin{equation}
    \forall h \forall m \in M(\mathcal{G}) \exists m_h \in M(\mathcal{G}_h): m_h \subset m.
\end{equation}
\end{manualtheorem}
\begin{proof}
If the threshold is lower than any score $s^p_i$, then $\mathcal{G}_h = \mathcal{G}$, and $m_h = m$ satisfies the theorem. Otherwise, some predictions are filtered by the threshold and $\mathcal{G}_h \subset \mathcal{G}$.

Without the loss of generality, assume that the threshold is low enough to filter out only one prediction. Otherwise, we can construct a chain of thresholds $h_1 < h_2 < \dots < h$, with each subsequent threshold filtering an additional vertex, and apply the theorem iteratively.

Denote $i$ to be the only vertex present in $\mathcal{G}$ and filtered from $\mathcal{G}_h$. If there is no edge containing vertex $i$ in the matching $m(\mathcal{G})$, then $m(\mathcal{G})$ is also the optimal matching in $\mathcal{G}_h$ and theorem holds. 

Consider the case when matching $m(\mathcal{G})$ contains an edge $(i, j)$. Let $\hat{m}_h = m(\mathcal{G}) \setminus \{(i, j)\}$, i.e., the matching without edge $(i, j)$. If it is optimal for the graph $\mathcal{G}_h$, then it will satisfy the theorem. Otherwise, there is an optimal matching $m_h$ with either (a) more vertices or (b) a smaller total weight than $\hat{m}_h$.

In (a), matching $m_h$ has a size greater than the size of $\hat{m}_h$. It follows that $m_h$ has the maximum size in the complete graph $\mathcal{G}$. If the total weight of $m_h$ is larger or equal to the optimal weight in $\mathcal{G}$, then it can not be less than the total weight of $\hat{m}_h$, which contradicts our assumption. At the same time, the total weight of $m_h$ can not be less than the optimal weight in the complete graph $\mathcal{G}$. It follows that case (a) leads to a contradiction.

Consider the case (b) when the size of $m_h$ equals $\hat{m}_h$. If $m_h$ does not include vertex $j$, then both $m_h$ and $\hat{m}_h$ are optimal. Otherwise, we can remove vertex $j$ from $\mathcal{G}_h$ and construct a new optimal matching $m'_h$ without $j$ by using Lemma \ref{theorem:lemma}, and again, both matchings $m'_h$ and $\hat{m}_h$ are optimal. This concludes the proof.
\end{proof}

{\bf Algorithm.} Using the theorem, we can introduce an effective algorithm for T-mAP computation. T-mAP is defined on a batch of predictions $\{\mathcal{S}_p^i\}_{i=1}^{n}$ and ground truth sequences $\{\mathcal{S}_{gt}^i\}_{i=1}^{n}$. Let $\mathcal{S}^{i,l}_{gt}$ denote a subsequence of the ground truth sequence $\mathcal{S}^i_{gt}$ containing all events with label $l$. By definition, multiclass T-mAP is the average of the average precision (AP) values for each label $l$:
\begin{equation}
    \mathrm{T\text{-}mAP}(\{\mathcal{S}_p^i\}, \{\mathcal{S}_{gt}^i\}) = \frac{1}{L} \sum\limits_{l=1}^L\mathrm{AP}(\{\mathcal{S}_p^i\}, \{\mathcal{S}_{gt}^{i, l}\}).
\end{equation}
Consider AP computation for a particular label $l$. AP is computed as the area under the precision-recall curve:
\begin{equation}
    \mathrm{AP} = \sum\limits_i \left(\mathrm{Rec}_i - \mathrm{Rec}_{i - 1}\right)\mathrm{Prec}_i,
\end{equation}
where $\mathrm{Rec}_i$ is the $i$-th recall value in a sorted sequence and $\mathrm{Prec}$ is the corresponding precision. Iteration is done among all distinct recall values ($\mathrm{Rec}_0 = 0$).

We have several correct and incorrect predictions for each threshold $h$ on the predicted label probability. These quantities define precision, recall, and the total number of ground truth events. A prediction is correct if it has an assigned ground truth event within the required time interval $|t^p_i - t_j^{gt}| \le \delta$. Note that each target can be assigned to at most one prediction. Therefore, we define a matching, i.e., the correspondence between predictions and ground truth events. The theorem states that the maximum size matching for each threshold $h$ can be constructed as a subset of an optimal matching between full sequences $\{\mathcal{S}_p^i\}_{i=1}^{n}$ and $\mathcal{S}^{i,l}_{gt}$, where optimal matching is a solution of the assignment problem with a bipartite graph, defined in Section \ref{sec:map-computation}.

As the matching, without loss of generality, can be considered constant for different thresholds, we can split all predictions into two parts: those that were assigned a ground truth and those that were unmatched. The matched set constitutes potential true or false positives, depending on the threshold. The unmatched set is always considered a false positive. Similarly, unmatched ground truth events are always considered false negatives. The resulting algorithm for AP computation for each label $l$ involves the following steps:
\begin{enumerate}
    \item Compute the optimal matching between predictions and ground truth events.
    \item Collect (a) scores of matched predictions, (b) scores of unmatched predictions, and (c) the number of unmatched ground truth events.
    \item \label{item:binary-problem}Assign a positive label to matched predictions and a negative label to unmatched ones.
    \item Evaluate maximum recall as the fraction of matched ground truth events.
    \item Find the area under the precision-recall curve for the constructed binary classification problem from item \ref{item:binary-problem} and multiply it by the maximum recall value.
\end{enumerate}
We have thus defined all necessary steps for T-mAP evaluation. Its complexity is $\mathcal{O}(LBN^3)$, where $L$ is the number of classes, $B$ is the number of sequence pairs, and $N$ is the maximum length of predicted and ground truth sequences.

{\bf Calibration dependency.} Calibration influences the weights assigned to the edges of the graph $\mathcal{G}$. T-mAP computation involves two key steps for each class label: matching and AP estimation. While average precision (AP) is invariant to monotonic transformations of predicted class logits, the matching step is only invariant to linear transformations. Specifically, the optimal matching seeks to minimize the total weight in the following form:
\begin{equation}
    C = \argmin\limits_{m \in M(\mathcal{G})} \sum\limits_{(i,j) \in m} (-s^p_i).
\end{equation}
A linear transformation of logits with a positive scaling factor will adjust the total weight accordingly, but the optimal matching will remain unchanged. Since the matching is performed independently for each class, we conclude that T-mAP is invariant to linear calibration.



\section{T-mAP Hyperparameters}
\label{app:tmap-hopt-full}
\begin{figure*}[h]
\centering
  \includegraphics[width=\textwidth]{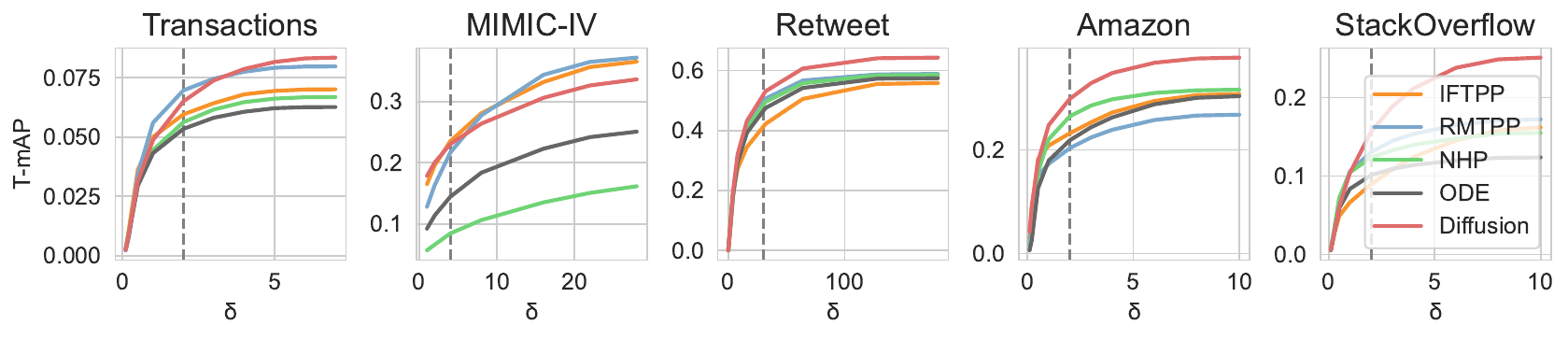}
  \caption{T-mAP dependency on the $\delta$ parameter. The dashed line indicates the selected value.}
  \label{fig:tmap-hyper-full}
  \vspace{-0.1in}
  
\end{figure*}

The dependency of T-mAP on the $\delta$ parameter for different datasets is presented in Figure~\ref{fig:tmap-hyper-full}.

\section{T-mAP for Highly Irregular Sequences}
\label{app:irregular}
\begin{figure*}[h]
\centering
  \includegraphics[width=.85\textwidth]{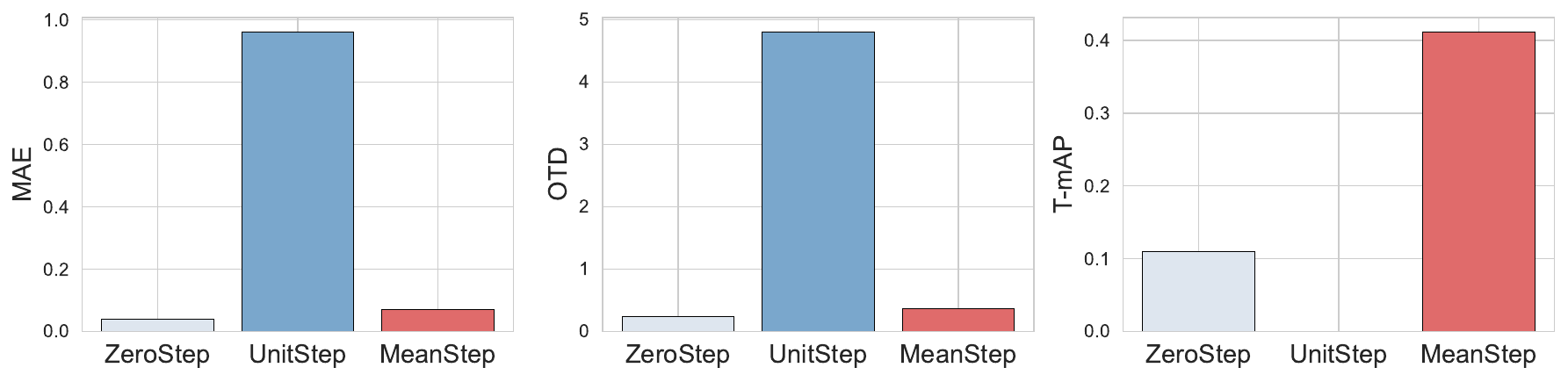}
  \caption{Comparison of simple baselines on the Toy dataset with highly irregular time intervals. MAE and OTD metrics represent error values, while T-mAP measures model quality.}
  \label{fig:toy-irregular}
  \vspace{-0.2in}
\end{figure*}

This section analyzes a toy dataset containing highly irregular event sequences. This dataset includes a single label, which aims to predict timestamps. Most time intervals in the dataset are zero, with only 5\% of intervals equal to one. We compare three simple baselines:
\begin{itemize}
    \item {\bf ZeroStep}, which predicts events with timestamps identical to the last observed event (zero intervals);
    \item {\bf UnitStep}, which predicts events with a unit time step (the largest time step in the dataset);
    \item {\bf MeanStep} predicts events using the average time step computed from historical data.
\end{itemize}
Evaluation results are shown in Fig.~\ref{fig:toy-irregular}. The results indicate that the MAE and OTD metrics assign the lowest error to the ZeroStep baseline, which simply repeats the last event without accounting for the dataset's irregularity. In contrast, T-mAP identifies the MeanStep baseline as the most effective, as it is the only method that analyzes historical data and incorporates timestamp statistics (mean interval) into its predictions.

These findings suggest that T-mAP is a more appropriate metric for assessing the ability of methods to predict irregular events.

\section{T-mAP for Long-tail Prediction Problems}
\label{app:long-tail}
\begin{figure*}[h]
\centering
  \includegraphics[width=\textwidth]{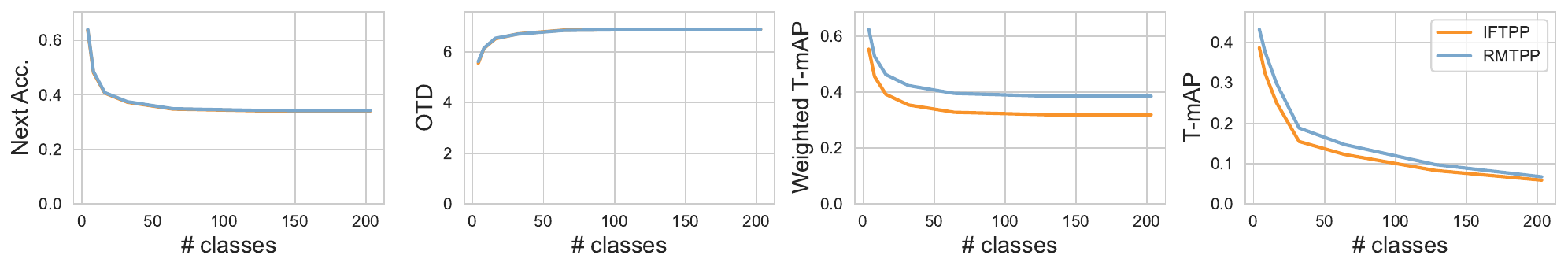}
  \caption{Comparison of various metrics on the Transactions dataset across subsets with fewer event classes.}
  \label{fig:long-tail}
\end{figure*}

In this section, we assess the capability of various evaluation metrics to capture long-tail prediction quality, specifically the ability of models to predict rare classes. Unlike OTD, T-mAP evaluates each class independently, allowing for different aggregation strategies. The standard T-mAP computes a macro average, where the quality for each event class contributes equally to the final score. Additionally, the HoTPP benchmark includes a weighted variant of T-mAP, where the weights are proportional to class frequencies. Fig.~\ref{fig:long-tail} compares the performance of IFTPP and RMTPP on the Transactions dataset, which includes 203 classes. The results show that all metrics, except macro T-mAP, remain unaffected as the dataset size increases beyond 60 classes. In contrast, macro T-mAP effectively evaluates the ability of models to predict across all available classes, highlighting its suitability for long-tail prediction tasks.

\section{HoTPP Benchmark Details}
\label{app:benchmark}
\subsection{Architecture}
\begin{figure*}[p]
\centering
  \includegraphics[width=\textwidth]{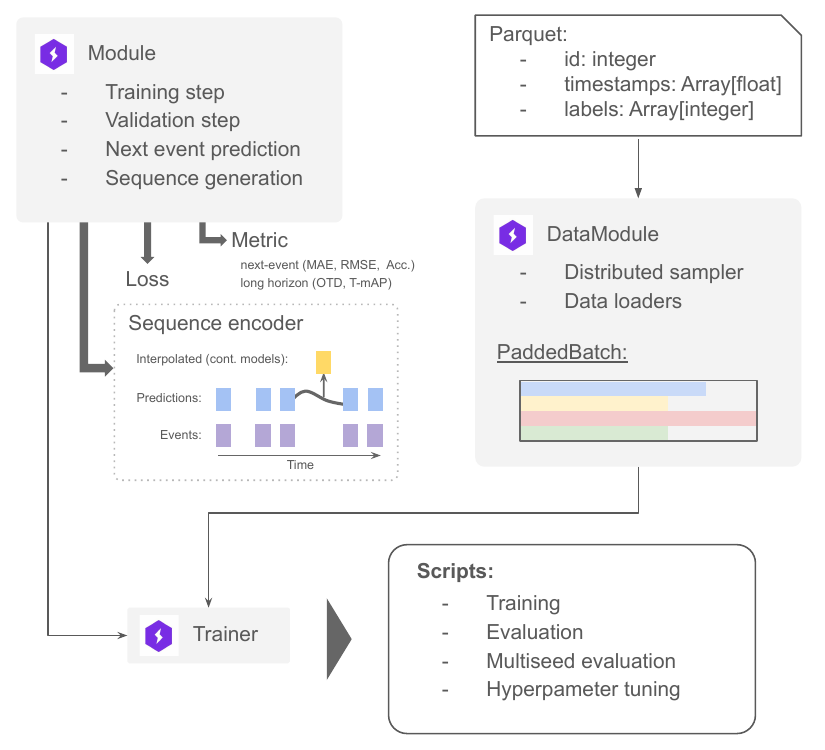}
  \caption{The architecture of the HoTPP library.}
  \label{fig:hotpp-arch}
\end{figure*}

HoTPP incorporates best practices from extensible and reproducible ML pipelines. It leverages PyTorch Lightning \cite{Falcon_PyTorch_Lightning_2019} as the core training library, ensuring reproducibility and portability across various computing environments. Additionally, HoTPP utilizes the Hydra configuration library \cite{Yadan2019Hydra} to enhance extensibility. The overall architecture is illustrated in Fig.~\ref{fig:hotpp-arch}. HoTPP supports both discrete-time and continuous-time models as well as RNN and Transformer architectures. Implementing a new method requires only adding essential components, while the rest can be configured through Hydra and configuration files.

\subsection{Metrics}
MTPP models are typically evaluated based on the accuracy of next-event predictions. Common metrics include mean absolute error (MAE) or root mean square error (RMSE) for time shifts and accuracy or error rate for label predictions. Some studies also assess test set likelihood as predicted by the model, but we do not include this measure as it is intractable for the IFTPP and Diffusion models. Previous works have advanced long-horizon evaluation using OTD~\cite{mei2019imputing, xue2023easytpp}. In addition to these metrics, we evaluate the novel T-mAP metric, which addresses the shortcomings of previous metrics, as discussed in Section \ref{sec:otd-limitations}.

\subsection{Backbones}
We use three types of architectures in our experiments. The first is the GRU network \cite{cho2014gru}, one of the top-performing neural architectures for event sequences \cite{babaev2022coles}. We also implement the continuous-time LSTM (CT-LSTM) from the NHP method \cite{mei2017nhp}. Since CT-LSTM requires a specialized loss function and increases training time, we use it exclusively with the NHP method, preferring GRU in other cases. An additional advantage of GRU is that its output is equal to its hidden state, simplifying the estimation of intermediate hidden states for autoregressive inference starting from the middle of a sequence. Lastly, we implement the AttNHP continuous-time transformer model~\cite{yang2022anhp}.

\subsection{Datasets}
\label{app:datasets}

Transactions\footnote{\url{https://huggingface.co/datasets/dllllb/age-group-prediction}}, Retweet\footnote{\url{https://huggingface.co/datasets/easytpp/retweet}}, Amazon\footnote{\url{https://huggingface.co/datasets/easytpp/amazon}}, and StackOverflow\footnote{\url{https://huggingface.co/datasets/easytpp/stackoverflow}} datasets were obtained from the HuggingFace repository. Transactions data was released in competition and came with a free license\footnote{\url{https://www.kaggle.com/competitions/clients-age-group/data}}. Retweet, Amazon, and StackOverflow come with an Apache 2.0 license. MIMIC-IV is subject to PhysioNet Credentialed Health Data License 1.5.0, which requires ethical use of this dataset. Because of a complex data structure, we implement a custom preprocessing pipeline for the MIMIC-IV dataset.

{\bf Notes on MIMIC-IV preprocessing}

MIMIC-IV is a publicly available electronic health record database that includes patient diagnoses, lab measurements, procedures, and treatments. We leverage the data preprocessing pipeline from EventStreamGPT to construct intermediate representations in EFGPT format. This process yields three key entities: a subjects data frame containing time-independent patient records, an events data frame listing event types occurring to subjects at specific timestamps, and a measurements data frame with time-dependent measurement values linked to the events data frame.

We combine events and measurements to create sequences of events for each subject. The classification labels are generally divided into two categories: diagnoses, represented by ICD codes, and event types, such as admissions, procedures, and measurements. A key challenge is that relying solely on diagnoses results in sequences that are too short, while using only event types leads to highly repetitive sequences dominated by periodic events, such as treatment start and finish.

However, using both diagnoses and event types together introduces another issue: diagnoses are sparsely distributed within a constant stream of repeated procedures, leading to imbalanced class distributions and poor performance. To mitigate this, we filter the data by removing duplicate events between diagnoses, allowing us to retain useful treatment data while preserving class balance.

The final labels are created by converting ICD-9 and ICD-10 codes to ICD-10 chapters using General Equivalence Mapping (GEM) and adding event types as additional classes. This conversion is necessary because the number of ICD codes is too large to use them directly as classes. Additionally, we address the issue of multiple diagnoses occurring in a single event by sorting timestamps for reproducibility.

\section{Performance improvements}
\label{app:performance-improvements}
The HoTPP benchmark provides highly optimized training and inference procedures for the efficient evaluation of datasets containing up to tens of millions of events. First, we implement parallel RNN inference, which reuses computations when inference is initiated from multiple starting points within a sequence. Additionally, we optimize the code and apply PyTorch Script to the CT-LSTM model from NHP and the ODE model. CT-LSTM from EasyTPP and the ODE model from the original repository serve as baselines. The timing results are presented in Table \ref{tab:hotpp-timings}. Experiments were conducted using a synthetic batch with a size of 64, a sequence length of 100, and an embedding dimension of 64. Evaluation was performed on a single Nvidia RTX 4060 GPU. The results show that HoTPP is 17 times faster at RNN autoregressive inference compared to simple prefix extension. HoTPP also accelerates CT-LSTM and ODE by 4 to 8 times. These optimizations significantly extend the applicability of the implemented methods to larger-scale datasets\footnote{Experiments are documented in the ``notebooks'' folder of the HoTPP repository}. Additionally, HoTPP offers a GPU implementation of the Hungarian algorithm, which also finds applications in computer vision.

\begin{table*}[h]
\vspace{0.2in}
\caption{HoTPP computation speed improvements in terms of seconds per batch.}
\centering
\begin{tabular}{l|ccccc}
\toprule
\multirow{4}{*}{Implementation} & \multicolumn{5}{c}{Study} \\
\cmidrule{2-6}
& \thead{RNN autoreg. \\ inference} & \thead{CT-LSTM \\ inference} & \thead{CT-LSTM \\ train} & \thead{ODE \\ inference} & \thead{ODE \\ train} \\
\midrule
Baseline & 2.44 & 0.0158 & 0.0519 & 0.08 & 0.162 \\
\bf HoTPP & \bf 0.14 & \bf 0.0025 & \bf 0.0276 & \bf 0.01 & \bf 0.048 \\
\hline
\textit{Impr.} & 17x & 6x & 2x & 8x & 3x \\
\bottomrule
\end{tabular}
\vspace{0.1in}
\label{tab:hotpp-timings}
\end{table*}

\section{Datasets Analysis}
\label{app:domain-analysis}
HoTPP incorporates datasets from various domains, including finance, social networks, and healthcare. Below, we provide additional details highlighting the key differences among these datasets.

\paragraph{Ordering Sensitivity.} While most datasets—such as StackOverflow, Amazon, Retweet, and Transactions—exhibit a relatively balanced distribution of timestamps, we observed that in MIMIC-IV, the proportion of zero time steps exceeds 50\%, as shown in Table \ref{tab:timestep-percentilles}. This results in a significant number of events sharing the same timestamp. Consequently, the actual order of events within the dataset may hold greater significance during evaluation than the precise prediction of timestamps. This characteristic can lead to instability in timestamp-based metrics, such as OTD and T-mAP, compared to index-based metrics like next-event quality, pairwise MAE and accuracy. The low next-event accuracy of methods based on NHP loss (NHP, AttNHP, and ODE) on MIMIC-IV can be attributed to their independent modeling of each event class, which results in a random ordering of events with identical timestamps.

\begin{table*}[h]
\caption{Time step percentiles.}
\centering
\begin{tabular}{l|ccccccc}
\toprule
Dataset& 1\% & 5\% & 10\% & 50\% & 90\% & 95\% & 99\% \\
\midrule
Transactions & 0.0 & 0.0 & 0.0 & 1.0 & 2.0 & 2.0 & 5.0 \\
MIMIC-IV & 0.0 & 0.0 & 0.0& 0.0 & 6.9 & 63.8 & 768.0 \\
Retweet & 0.0 & 0.0 &1.0& 8.0& 85.0&151.0& 377.0 \\
Amazon & 0.01 & 0.01 & 0.01 & 0.73 & 0.79 & 0.79 & 0.80 \\
StackOverflow & 0.0003 & 0.01 & 0.05 & 0.51 & 2.16 & 2.97 & 5.08 \\
\bottomrule
\end{tabular}
\label{tab:timestep-percentilles}
\end{table*}

\section{Training details}
\label{app:training}
We trained each model using the Adam optimizer, which has a learning rate 0.001, and a scheduler that reduces the learning rate by 20\% after every five epochs. Gradient clipping was applied, and the maximum L2-norm was set to 1.

We performed computations on NVIDIA V100 and A100 GPUs, with some smaller experiments conducted on an Nvidia RTX 4060. Each method was trained on a single GPU. The training time varied depending on the dataset and method, ranging from 5 minutes for RMTPP on the StackOverflow dataset to 40 minutes for the same method on the Transactions dataset and up to 15 hours for NHP on Transactions. Multi-seed evaluations took approximately five times longer to complete.

\section{Hyperparameters}
\label{app:hyperparameters}

Dataset-specific training parameters are listed in Table \ref{tab:training-parameters}:

\begin{table*}[h]
\caption{Training hyperparameters.}
\begin{adjustbox}{width=\textwidth}
\begin{tabular}{l|c|c|c|c|c}
\toprule
Dataset& Num epochs & Max Seq. Len. & Label Emb. Size & Hidden Size & Head hiddens \\
\midrule
Transactions & 30 & 1200 & 256 & 512 & 512 $\rightarrow$ 256 \\
MIMIC-IV & 30 & 64 & 16 & 64 & 64 \\
Retweet & 30 & 264 & 16 & 64 & 64 \\
Amazon & 60 & 94 & 32 & 64 & 64 \\
StackOverflow & 60 & 101 & 32 & 64 & 64 \\
\bottomrule
\end{tabular}
\end{adjustbox}
\label{tab:training-parameters}
\end{table*}

\paragraph{Discrete Timestamps.} Some datasets feature continuous timestamps (e.g., StackOverflow, Amazon, MIMIC-IV), while others round timestamps to a specific precision (e.g., Retweet, Transactions). Modeling discrete timestamps presents a unique challenge, as density estimation methods like NHP can produce infinite PDF values. We introduce small Gaussian noise to discrete timestamps to address this issue, effectively smoothing the distribution. The degree of smoothing was carefully tuned for each dataset individually. The exact values are provided in the configuration files included with the source code.

\section{Predictions Diversity}
\label{app:qualitative-diversity}
\begin{figure*}[t]
\centering
  \includegraphics[width=\textwidth]{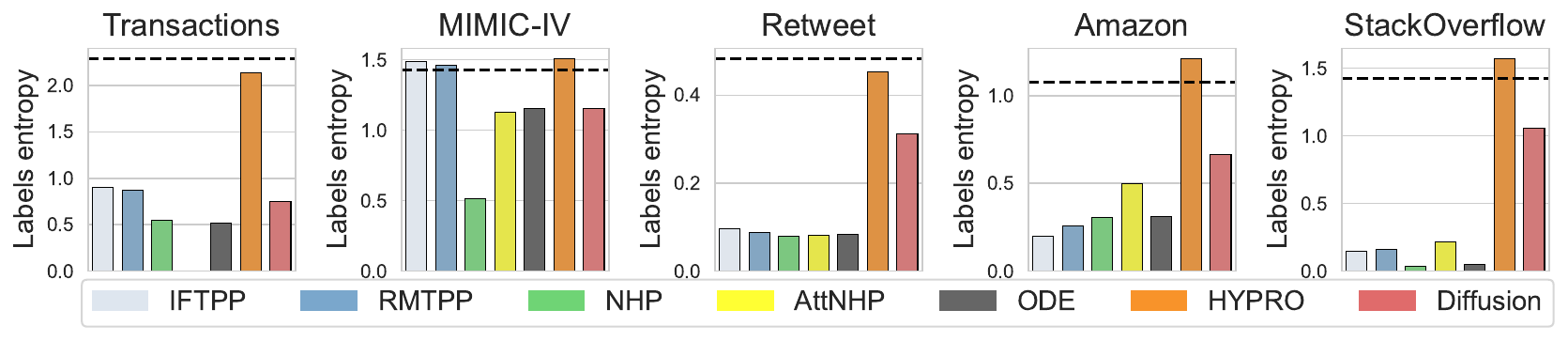}
  \caption{Comparison of predicted label entropies, with the dashed horizontal line indicating the ground truth entropy.}
  \label{fig:entropies-full}
\end{figure*}
Table \ref{tab:example-predictions} presents example predictions from various methods on the Transactions, MIMIC-IV, and Retweet datasets. The table focuses exclusively on predicted label sequences. It can be seen that all methods, except HYPRO, exhibit issues with constant or repetitive outputs. We believe this behavior stems from a bias in the predictions toward the most frequent labels, especially in scenarios with high uncertainty. For the MIMIC-IV dataset, the prediction patterns differ significantly, as most events follow a predefined sequence, such as admission, a standard set of laboratory tests, and diagnosis. In this case, the uncertainty is lower, enabling the methods to generate sequences with more diverse event types. Addressing the mentioned limitation in future research could lead to the development of methods capable of producing more varied and realistic sequences in high-uncertainty scenarios.

The quantitative results for all 5 datasets are presented in Figure~\ref{fig:entropies-full}. For details, please refer to Section~\ref{sec:collapse}.

\begin{table*}[h]
\caption{Example predictions (labels only).}
\centering
\begin{adjustbox}{width=\textwidth}
\begin{tabular}{l|c|ccc}
\toprule
Method & Seq. ID & Transactions & MIMIC-IV & Retweet \\
\midrule
\multirow{3}{*}{IFTPP}
& 0 & 3, 1, 3, 1, 3, 1, 3, 3, 3 & 10, 27, 23, 22, 1, 27, 3, 28, 26, 25, 23, 30 & 1, 1 \\
& 1 & 6, 3, 6, 6, 6, 6, 6 & 11, 10 & 0, 0, 0, 0, 0, 0, 0, 0, 0 \\
& 2 & 3, 1, 3, 1, 3, 1, 3, 1, 3 & 2, 7, 14, 12, 4, 6, 11, 10 & 1, 1, 1, 1, 1, 1, 1, 1, 1, 1, 1, 1 \\
\midrule
\multirow{3}{*}{RMTPP}
& 0 & 3, 1, 3, 1, 3, 1 & 10, 27, 23, 22, 1, 30, 27, 5, 8, 9, 16, 15 & 1, 1, 1, 1 \\
& 1 & 6, 6, 6, 6, 6, 6 & 11, 10, 1 & 0, 0, 0, 0, 0, 0, 0, 0 \\
& 2 & 3, 1, 3, 1, 3 & 2, 7, 14, 12, 4, 6, 11, 10 & 1, 1, 1, 1, 1, 1, 1, 1, 1, 1, 1, 1 \\
\midrule
\multirow{3}{*}{NHP}
& 0 & 1, 1, 1, 1, 1, 3, 1, 1, 1, 1 & 6 & 1, 1, 1, 1 \\
& 1 & 1, 1, 6, 6, 6, 6, 6 & 6, 6, 6 & 1, 1, 1, 1, 1, 1, 1 \\
& 2 & 1, 1, 1, 1, 1, 1, 1, 1, 1, 1 & 6, 6 & 1, 1, 1, 1, 1, 1, 1, 1, 1, 1, 1, 1 \\
\midrule
\multirow{3}{*}{ODE}
& 0 & 1, 1, 1, 1, 1, 1, 1, 1, 1, 1, 1 & 10, 1, 3, 5, 2, 4, 6, 1 & 1, 1, 1, 1 \\
& 1 & 6, 6, 6, 6, 6 & 1, 3, 5, 2, 4, 6, 1, 3, 5, 2, 4, 6 & 0, 0, 0, 0, 0, 0, 0 \\
& 2 & 1, 1, 1, 1, 1, 1, 1 & 2, 4, 6, 1, 3, 5, 2, 4, 10 & 1, 1, 1, 1, 1, 1, 1, 1, 1, 1, 1, 1 \\
\midrule
\multirow{3}{*}{HYPRO}
& 0 & 3, 1, 16, 3, 12, 1 & 23, 27, 22, 1, 28, 26, 25, 23 & 0, 0, 1, 0 \\
& 1 & 1, 32, 6, 6, 1, 6 & 1, 28, 25, 23, 3, 26, 22, 5 & 0, 0, 0, 1, 1, 0, 1 \\
& 2 & 3, 5, 1, 3, 5 & 15, 13, 2, 7, 4, 1, 3, 5 & 0, 0, 0, 0, 0, 0, 1, 0, 0, 0, 0, 0 \\
\midrule
\multirow{3}{*}{Diffusion}
& 0 & 1, 3, 1, 1, 1, 3, 18, 1 & 0, 0, 0, 0, 0, 0, 0, 0, 0, 0, 0, 0 & 0, 0, 0, 0, 0, 0, 0, 0, 0, 0, 0, 0 \\
& 1 & 6, 6, 6, 6, 6, 6, 6 & 0, 0, 0, 0, 0, 0, 0, 0, 0, 0, 0, 0 & 1, 1, 1, 0, 0, 1, 1, 1, 1, 0, 1, 0 \\
& 2 & 8, 1, 3, 8, 3, 8, 1, 3, 8, 1, 1, 8 & 0, 0, 0, 0, 0, 0, 0, 0, 0, 0, 0, 0 & 1, 1, 1, 1, 1, 0, 1, 1, 1, 1, 1, 1 \\
\bottomrule
\end{tabular}
\end{adjustbox}
\label{tab:example-predictions}
\end{table*}

\end{document}